\documentclass[10pt,twocolumn,letterpaper]{article}

\usepackage{cvpr}      % To produce the REVIEW version
\usepackage{fontawesome}
\usepackage{multirow}
\usepackage{arydshln}
\usepackage{comment}
\usepackage{subcaption}

\renewcommand{\paragraph}[1]{\noindent\textbf{#1}~}
\usepackage[dvipsnames]{xcolor}

\definecolor{cvprblue}{rgb}{0.21,0.49,0.74}
\usepackage[pagebackref,breaklinks,colorlinks,citecolor=cvprblue]{hyperref}

\def\paperID{968} % *** Enter the Paper ID here
\def\confName{CVPR}
\def\confYear{2024}

\title{Q-Instruct: Improving Low-level Visual Abilities\\ for Multi-modality Foundation Models}

\author{Haoning Wu$^{1\heartsuit}$, Zicheng Zhang$^{2\heartsuit}$, Erli Zhang$^{1\heartsuit}$,\\ Chaofeng Chen$^1$, Liang Liao$^1$, Annan Wang$^1$, Kaixin Xu$^4$, Chunyi Li$^2$,  Jingwen Hou$^1$,\\ Guangtao Zhai$^2$, Geng Xue$^4$, Wenxiu Sun$^3$, Qiong Yan$^3$,  Weisi Lin$^{1\diamondsuit}$ \\
$^1$Nanyang Technological University, $^2$Shanghai Jiaotong University, $^3$Sensetime Research, $^4$A*STAR
}

\begin{document}

\twocolumn[{%
\renewcommand\twocolumn[1][]{#1}%
\maketitle
\begin{center}
    \centering
    \vspace{-2.3em}
    \includegraphics[width=0.93\linewidth]{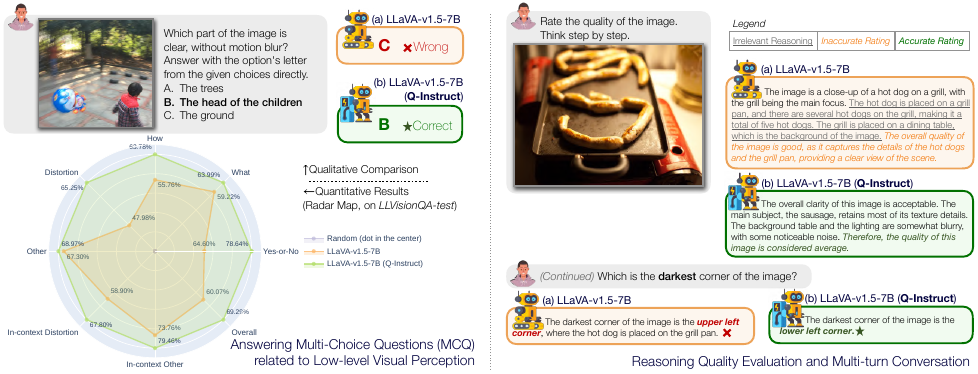}
    \vspace{-10pt}
    \captionof{figure}{Abilities of \textbf{Q-Instruct}-\textit{tuned} LLaVA-v1.5-7B~\cite{improvedllava} on various low-level visual tasks, in comparison with the baseline version.} \label{fig:one}
    
\end{center}%
}]
\vspace{1mm}

\maketitle

\let\thefootnote\relax\footnotetext{\noindent$^\heartsuit$Equal contribution. $^\diamondsuit$Corresponding author.}
\let\thefootnote\relax\footnotetext{$^\clubsuit$Project Page: \href{https://q-future.github.io/Q-Instruct}{\textit{https://q-future.github.io/Q-Instruct}}}

\begin{abstract}
 Multi-modality large language models (MLLMs), as represented by GPT-4V, have introduced a paradigm shift for visual perception and understanding tasks, that a variety of abilities can be achieved within one foundation model. While current MLLMs demonstrate primary \textbf{low-level visual abilities} from the identification of low-level visual attributes (e.g., clarity, brightness) to the evaluation on image quality, there's still an imperative to further improve the accuracy of MLLMs to substantially alleviate human burdens. To address this, we collect the first dataset consisting of human natural language feedback on low-level vision. Each feedback offers a comprehensive description of an image's low-level visual attributes, culminating in an overall quality assessment. The constructed \textbf{Q-Pathway} dataset includes 58K detailed human feedbacks on 18,973 multi-sourced images with diverse low-level appearance. To ensure MLLMs can adeptly handle diverse queries, we further propose a GPT-participated transformation to convert these feedbacks into a rich set of 200K instruction-response pairs, termed \textbf{Q-Instruct}. Experimental results indicate that the \textbf{Q-Instruct} consistently elevates various low-level visual capabilities across multiple base models. We anticipate that our datasets can pave the way for a future that foundation models can assist humans on low-level visual tasks. %Our code, dataset, and pre-trained weights are all released in \href{https://q-future.github.io/Q-Instruct}{\textit{https://q-future.github.io/Q-Instruct}}. 
 
 %To enhance these models, we conduct a large-scale subjective experiment collecting a vast number of human language feedbacks. Each feedback first describes the low-level quality-related perception about an image, and then conclude as an overall quality opinion, denoted as a \textbf{pathway quality evaluation}. Specifically, the constructed \textbf{Pathway-Q} database contains 18,973 images, each with at least three pathway quality evaluations. Furthermore, we design a GPT-assisted protocol to convert these pathway explanations into 110K diverse-concept and diverse-format instruction-response pairs on these images, denoted as \textbf{Q-Instruct}. Experimental results indicate that Q-Instruct consistently elevates perception and evaluation abilities across several foundational models. We aspire for our datasets to pave the way for a general-purpose model that mirrors human-like quality perception and evaluation in the future.

\end{abstract}

\section{Introduction}
\label{sec:intro}

%

%\begin{figure}[t]
%  \centering
%  \includegraphics[width=0.9\linewidth]{imgs/TeaserQInstruct.pdf}
%   \vspace{-5pt}
%   \caption{LLaVA-v1.5-7B~\cite{improvedllava}, \textit{before} / \textit{after} low-level visual instruction tuning on the \textbf{Q-Instruct} with 200K instruction-response pairs, bringing notable improvements on low-level visual abilities. }
%  \label{fig:one}
%   \vspace{-12pt}
%\end{figure}

\begin{figure*}[t]
  \centering
  \includegraphics[width=0.99\linewidth]{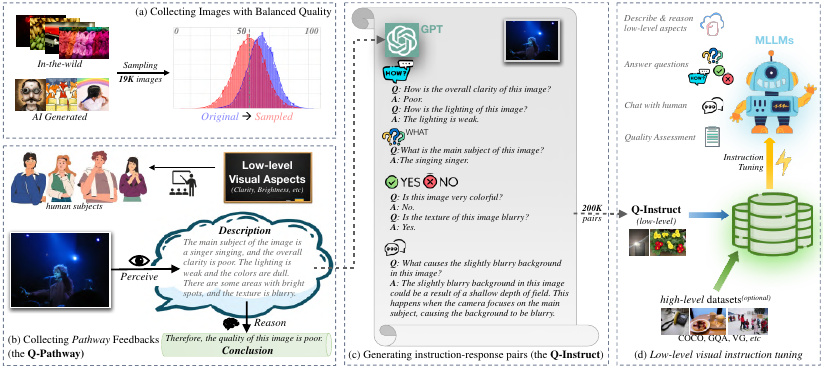}
    \vspace{-8pt}
   \caption{Data construction pipeline. First, we collect \textbf{58K} human feedbacks on low-level visual aspects (the \textbf{Q-pathway}, a/b); they are then converted into with \textbf{200K} instruction-response pairs (the \textbf{Q-Instruct}, c), which are used for (d) low-level visual instruction tuning.}
   \label{fig:two}
   \vspace{-12pt}
\end{figure*}
%-------------------------------------------------------------------------
Computer vision has witnessed a recent paradigm shift attributed to the emergence of multi-modality large language models (MLLMs)~\cite{openai2023gpt4,llava,iblip,llamaadapterv2}. These models aim to transcend traditional task-specific experts, and serve as general-purpose foundation models capable of facilitating humans across a variety of visual tasks~\cite{mllmsurvey}. Specifically, these foundation models also bring exciting potentials in the domain of \textbf{low-level visual perception and understanding}. This domain includes not only commonly-focused image quality assessment (IQA)~\cite{KonIq,paq2piq,wu2023dover} tasks, but also finer-grained abilities to identify the low-level visual attributes (\textit{noise, blur, etc})~\cite{koniqplusplus}, or evaluate the low-level visual dimensions (\textit{clarity, brightness, etc})~\cite{spaq,wu2023explainable}. As human cognition associated with these tasks is highly interconnected, we aspire for a unified foundation model to establish general abilities across these tasks, which could robustly respond to open-ended human queries on low-level visual aspects.

Nevertheless, though existing MLLMs can basically reply to human queries regarding low-level visual aspects, the accuracy of their responses remains unsatisfactory~\cite{mmbench,wu2023qbench} (Fig.~\ref{fig:one}(a)). The primary problem is the lack of low-level visual datasets during training MLLMs, where publicly available datasets generally only focus on high-level visual abilities~\cite{okvqa,cocovqa,gqa,vg}. To solve this problem, we construct the \textbf{Q-Instruct}, the first large-scale \textit{low-level visual instruction tuning} dataset, in the following two steps:

% 符合人的认知
\textit{Step 1: Collect human feedbacks for low-level vision.}

For this step, we invite human subjects to provide direct feedbacks on their low-level perception and understanding over a variety of images (Fig.~\ref{fig:two}(b)). Specifically, each feedback should include two parts: 1) Primarily, an exhaustive \textbf{description} on {elemental} low-level attributes (\textit{e.g.~blurs, noises, clarity, color, brightness}). Such descriptions should also include content~\cite{sfa,rfugc} or position~\cite{wu2022fastervqa,paq2piq} contexts (\textit{e.g.~\underline{the duck / the left part of the image} is under-exposed}) that are related to low-level attributes. 2) Then, an {overall} \textbf{conclusion} on the image quality based on the description of the attributes. With the two parts, the feedbacks, denoted as \textbf{\textit{pathway}} feedbacks, not only record fundamental human low-level perception but also reflect the human reasoning process on evaluating visual quality. The hence-constructed \textbf{Q-Pathway} dataset (Fig~\ref{fig:two}(b)) contains 58K pathway feedbacks on 18,973 multi-sourced images, each image with at least three feedbacks (\textit{avg. 46.4 words per feedback}).

\textit{Step 2: Convert these feedbacks for instruction tuning.}

While these \textit{pathway} feedbacks themselves make up an important subset for the \textit{low-level visual instruction tuning}, the full instruction tuning dataset should be designed to activate more capabilities. Primarily, it should also include a low-level \textit{visual question answering} (VQA) subset. To generate a reliable VQA subset, we refer to the setting that how COCO-VQA~\cite{cocovqa} is derived from image captions, and employ GPT~\cite{chatgpt} to convert the \textit{pathway} feedbacks into question-answer pairs with adjectives (\textit{e.g.~good/fair/poor}) or nouns (\textit{e.g.~noise/motion blur}) as answers. Similarly, we also collect a balanced \textit{yes-or-no} question-answer set based on the information in the feedbacks (\textit{answered with yes}), or information contrast to the feedbacks (\textit{answered with no}); some context-related question-answer pairs are also created to better ground~\cite{refcoco} the low-level attributes. Following existing studies~\cite{AOKVQA}, all question-answer pairs in the VQA subset include both multiple-choice (\textit{A/B/C/D}) and direct-answer settings. Furthermore, besides the VQA subset, with the assistance of GPT, we also collect a subset of long conversations related to the low-level concerns (\textit{e.g.~why the distortions happen}, \textit{how to improve the picture quality}). The subsets compose into the \textbf{Q-Instruct} dataset (Fig.~\ref{fig:two}(c)) with 200K instruction-response pairs, which is designed to enhance MLLMs on a variety of low-level visual abilities.

The core contributions of our study can be summarized as follows: \textbf{1)} We collect the \textbf{Q-Pathway}, a multi-modality dataset for low-level visual perception and quality assessment, which includes direct human feedbacks (\textit{with reasoning}) on low-level visual aspects. \textbf{2)} Based on \textbf{Q-Pathway}, we construct the \textbf{Q-Instruct}, the first instruction tuning dataset that focuses on human queries related to low-level vision. \textbf{3)} Our rich experiments on  \textit{low-level visual instruction tuning} ((Fig.~\ref{fig:two} (d)) validate that the \textbf{Q-Instruct} improve various low-level abilities of MLLMs (Fig.~\ref{fig:one}), and bring insights for future studies to inject various low-level visual abilities into the scope of general foundation models.

%to robustly describe the low-level visual attributes, as the low-level counterpart of image captioning~\cite{cococaps,flickrcaps,nocaps}. Moreover, we split the \textbf{elemental} and \textbf{overall} sentences into a two-round instruction-response conversation, so as to improve reasoning~\cite{cot} ability for low-level vision. 

%To improve the low-level visual abilities of MLLMs, a large-scale low-level visual instruction tuning dataset is indispensable. 

\section{Related Works}

\subsection{Low-level Visual Perception}
%Low-level visual perception is crucial for visual applications because it provides the foundational information that allows software and systems to interpret, analyze, and respond to visual stimuli. Image Quality Assessment (IQA) is an integral part of low-level visual perception, focusing on the foundational visual cues that lay the groundwork for advanced visual analysis. 
\paragraph{Tasks and Datasets.} % IQA Datasets, attribute-level IQA, 等等
 Image quality assessment (IQA), targeting to predict accurate scores aligned with integrated human opinions on all low-level aspects, has always been the chief task in low-level visual perception. Many datasets are developed to address IQA on artificially-distorted images~\cite{livemultipledistortions,kadid} (\textit{JPEG, AWGN, etc}), in-the-wild photographs~\cite{KonIq,paq2piq}, or recently-popular AI-generated contents~\cite{agiqa3k,imagereward}, providing important metrics for visual content production and distribution. Despite general IQA, recent studies have started to focus on finer-grained low-level visual aspects, and explored some related tasks such as evaluating on low-level visual dimensions (\textit{e.g.~color, brightness})~\cite{spaq,wu2023explainable}, or distinguishing the existing distortions (\textit{e.g.~blur, noise, over-exposure}) in images~\cite{koniqplusplus}. Some recent works~\cite{wu2023dover,wu2023bvqi,bvqiplus} also consider some photography-related dimensions (\textit{e.g.~composition, lighting, bokeh})~\cite{aadb} as a broader sense of low-level aspects. In general, low-level visual perceptual tasks can include all aspects of image appearance (\textit{in contrast to object-level contents}) that can be perceived by human and evoke different human feelings. While these low-level visual tasks used to be tackled separately, the proposed datasets bring the opportunities to include, relate and learn these tasks together, supporting one foundational model to generally master on these tasks.

 %Nevertheless, unlike IQA, with less applicable task-specific data, there still remains a gap for current machine learning algorithms to accurately and robustly perform this tasks.

%focusing on visual distortions such as noises and blurs~\cite{koniqplusplus,wu2023explainable}, alongside other low-level attributes like color, lighting, and composition~\cite{aadb}. These elements play a role in the aesthetics and emotions evoked by natural photographs~\cite{avaiaa}, as well as human preferences for images from computer graphics~\cite{zhang2023subjective} and AI sources~\cite{agiqa3k,imagereward}.

\paragraph{Approaches.} % NIQE 基于规则；deep learning based 基于task-specific数据；CLIP-IQA 基于V-L，good zero-shot，但是对attribute level不够准确；MLLM can also do, but not so good.
Similarly, the approaches designed for low-level visual perception also basically focus on their general IQA abilities. The traditional IQA metrics, \textit{e.g.}~NIQE~\cite{niqe}, operate on discipline-based methodologies without training with human opinions, offering robust but less accurate evaluations. In contrast, deep learning-based methods~\cite{dbcnn,hyperiqa,musiq,dists,topiq,wu2022fastvqa} utilize task-specific data, capitalizing on the extensive learning capacities of neural networks to tailor their assessment to particular data distributions, while they also suffer from compromised generalization abilities. Notably, recent methods~\cite{clipiqa,liqe,clipiaa,clip3dqa,vila} explore CLIP~\cite{CLIP} for IQA, which stand out for their pioneer efforts on \textbf{\textit{multi-modality integration}} for low-level vision, and exciting zero-shot performance. Their zero-shot IQA abilities are also inherited by most recent MLLMs~\cite{improvedllava,xcomposer,QWen-VL}. Similar as NIQE, these multi-modality IQA methods are robust on various scenarios, yet not enough accurate on each single case. While these methods present improving performance on general IQA, the other finer-grained low-level visual perception abilities are still yet to be deeply investigated; moreover, tackling all these tasks separately may overlook the underlying relationships between them, refraining from reasoning among these sections. After instruction tuning with the proposed \textbf{Q-Instruct}, MLLMs can significantly improve their abilities on various low-level visual abilities, forecasting a future to unify these tasks through one model.

%it has its limitations when it comes to the attribute level.

%\subsection{Multi-modality Foundation Models}
\subsection{Multi-modality Large Language Models} % CLIP multi-modality model; LLM + CLIP -> MLLM, 具体他们可以实现什么（high-level），low-level 上还不能做到什么。

Large language models (LLMs), \textit{e.g.} GPT-4~\cite{openai2023gpt4}, T5~\cite{flant5}, LLaMA~\cite{llama}, has shown great language abilities regarding general human knowledge. With CLIP~\cite{CLIP} and additional adapting modules to involve visual inputs into LLMs, the multi-modality large language models (MLLMs) \cite{otter,llamaadapterv2,llava,iblip,xcomposer} can tackle a variety of multi-modality tasks for high-level vision, such as \textit{image captioning}~\cite{cococaps,nocaps,flickrcaps}, \textit{visual question answering} (VQA)~\cite{cocovqa,okvqa,AOKVQA}, and more language-related capabilities~\cite{mmbench,mme,seedbench}. Nevertheless, the evaluation results in the recent benchmark~\cite{wu2023qbench} reveal that MLLMs' low-level visual abilities are still unsatisfactory, especially when it comes to the \textit{finer-grained} low-level perception questions. While we notice that this is mainly due to the lack of respective data, we collect the first \textit{low-level visual instruction tuning} dataset, the \textbf{Q-Instruct}, to improve low-level visual abilities for different MLLMs, and bring them into the realm of low-level visual perception.

%Building upon this innovation of CLIP~\cite{CLIP} (a cutting-edge multi-modal model bridging visual and language understanding in a unified framework), the advent of large language models (LLMs) such as ChatGPT, Bard, and their open-source counterparts like LLaMA~\cite{llama} and MPT~\cite{mpt}, has marked a significant shift towards general-purpose AI capabilities. Building on this momentum, MLLMs like LLaVA~\cite{llava}, MiniGPT-4~\cite{minigpt4}, InstructBLIP~\cite{iblip}, and Otter~\citep{otter} have further enhanced the visual AI domain. While they're adept at overarching how-level visual tasks and human interaction, pinpointing intricate low-level visual details remains a challenge.

%\subsection{Multi-modality Tasks for High-level Vision} % image captioning, visual question answering, 发展路径 

%While we aim to bring multi-modality (\textit{vision-language}) to boost a broad range of low-level vision tasks, the past efforts on high-level vision will be great experience for us.  
%In the realm of Vision-Language Tasks, the capabilities of MLLMs are actively explored and validated. They demonstrate proficiency not only in traditional vision tasks like image classification or segmentation~\cite{lai2023lisa} but also in integrated vision-language tasks. These encompass areas such as image captioning~\cite{cococaps}, visual question answering~\cite{cocovqa}, and cross-modality grounding~\cite{kosmos2}. This showcases the versatility and adaptability of MLLMs across both foundational and advanced vision-language domains.

\section{the \textit{Q-Pathway}}

%\subsection{Overview}

As the fundamental part of the dataset construction, we introduce the \textbf{Q-Pathway}, the first large scale dataset that collects \textbf{text} feedbacks from human on low-level visual aspects. To diversify and balance different low-level appearances, we sub-sample images from \textbf{seven} sources (Sec.~\ref{sec:31}) and reduce the \textbf{\textit{skews}} in the source distributions (Tab.~\ref{tab:1}). After the preparation of images, we discuss the rationality and the detailed task definition for the \textit{pathway} feedbacks (Sec.~\ref{sec:32}), a kind of natural language feedback, as collected in the \textbf{Q-Pathway}. The subjective study is conducted \textbf{in-lab} (Sec.~\ref{sec:33}), where all subjects are trained before providing feedback. The analysis of the \textbf{Q-Pathway} is in Sec.~\ref{sec:34}.

%The assembled \textbf{Q-Pathway} dataset comprises 58K comprehensive human responses related to 18,973 images sourced from multiple origins, showcasing a variety of low-level visual features, which is the first large-scale multi-modal dataset specifically for low-level visual attributes. 

\begin{table}[!t]\small
    \centering
\renewcommand\arraystretch{1.10}
\renewcommand\tabcolsep{4pt}
    \caption{The \textbf{Q-Pathway} compared to its sources. We sub-sample the source images to reduce the \textbf{\textit{skews}} in their $\mathrm{MOS}$ distributions, resulting in the sampled distribution to be further \underline{balanced}. }
    \vspace{-6pt}
   \resizebox{\linewidth}{!}{\begin{tabular}{l|ccc|ccc}
    \toprule
    \multirow{2}{80pt}{\textbf{Image Sources \textcolor{gray}{$\mathrm{MOS}\in[0,100)$}}} & \multicolumn{3}{c|}{Original Distribution} & \multicolumn{3}{c}{Sampled Distribution}  \\ \cline{2-7}
      & {Size} & {$\mu_{\mathrm{MOS}}$} & $\sigma_{\mathrm{MOS}}$ & {Size} & {$\mu_{\mathrm{MOS}}$} & $\sigma_{\mathrm{MOS}}$\\ \hline
     KonIQ-10k~\citep{KonIq} & 10,073 & 58.73  & 15.43  & 5,182 & 49.53 & 15.72\\
     SPAQ~\citep{spaq} & 11,125 & 50.32 & 20.90 & 10,797 & 49.46 & 20.63\\
     LIVE-FB~\citep{paq2piq} & 39,810 & 72.13 & 6.16 & 800 & 60.68 & 17.38  \\
     LIVE-itw~\citep{clive} & 1,169 & 55.38 & 20.27 & 200 & 55.70 & 19.83  \\ 
     AGIQA-3K~\citep{agiqa3k} & 2,982 & 50.00 & 19.80 & 400 & 40.80 & 21.80 \\
     ImageRewardDB~\citep{imagereward} & 50,000 & \multicolumn{2}{c|}{\textit{- w/o $\mathrm{MOS}$ -}} & 584 & \multicolumn{2}{c}{\textit{- w/o $\mathrm{MOS}$ -}}\\
     15\textit{-distortion} COCO~\citep{cococaps} & 330,000 &  \multicolumn{2}{c|}{\textit{- w/o $\mathrm{MOS}$ -}} & 1,012 & \multicolumn{2}{c}{\textit{- w/o $\mathrm{MOS}$ -}}   \\ \hline
     \textit{Overall}  & 445,159 & 65.02 & 16.51 & 18,973 & \underline{49.87} & \underline{19.08}  \\
    \bottomrule
\end{tabular}
 }
\vspace{-12pt}
    \label{tab:1}
\end{table}

\begin{figure*}
    \centering
    \includegraphics[width=\linewidth]{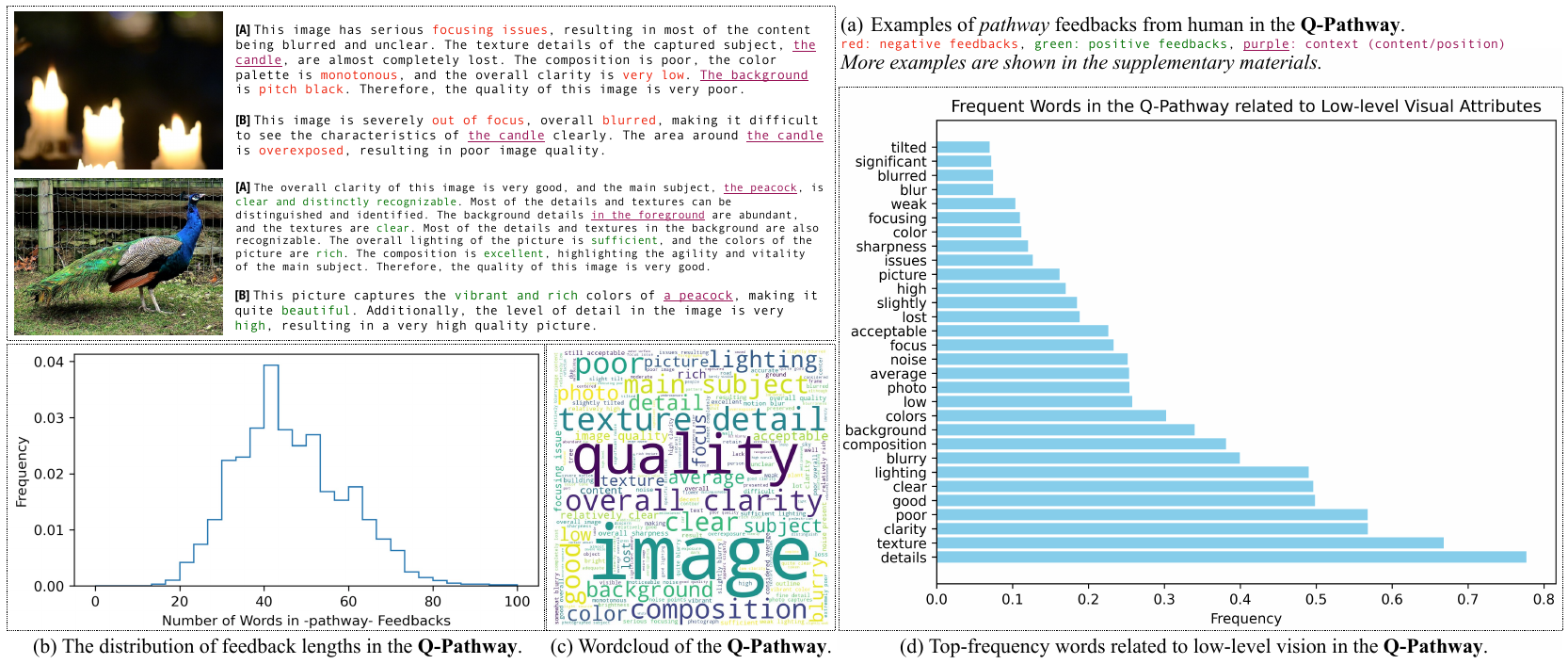}
    \caption{(a) Example \textit{pathway} feedbacks, each containing a detailed description followed by an overall evaluation, with context included. (b) The distribution of \textit{pathway} feedback lengths. (c) \textit{Wordcloud} of the \textbf{Q-Pathway}. (d) Top-frequency words related to low-level vision.}
    \label{fig:three} 
    \vspace{-10pt}
\end{figure*}

\subsection{Preparation of Images}
\label{sec:31}

The images in the \textbf{Q-Pathway} are sampled from various sources, including four \textit{in-the-wild} IQA datasets~\cite{spaq,KonIq,paq2piq,clive}, and two datasets with \textit{AI-generated} images~\cite{agiqa3k,imagereward}. Specifically, as compared in Tab.~\ref{tab:1}, the sub-sampled population of images is carefully constructed to introduce more diverse low-level appearances in the \textbf{Q-Pathway}, which is neither skewed towards positive appearances nor towards negative appearances.
%The two major sources of the \textbf{Q-Pathway}, SPAQ and KonIQ-10K, are noticed with slightly or skewed towards high-quality images (\textit{i.e.} with mean opinion score $\mathrm{MOS}>50$). While it has been observed~\cite{wu2023qbench} that base MLLMs are mostly prone towards positive answers, we would not like our collected dataset to increase such bias. Henceforth, we design an algorithm to remove a part of images with \textit{better-than-average} quality, resulting in a more balanced distribution between positive and negative low-level appearances. For another popular large-scale in-the-wild dataset, the LIVE-FB~\cite{paq2piq}, as its original distribution is far more skewed, we collect a smaller subset of it with significantly-reduced skewness. For AI-generated images, noticing that images in ImageRewardDB~\cite{imagereward} generally have good low-level appearance, we obtain \textit{worse-than-average} images from AGIQA-3K~\cite{agiqa} database a a compensation.
Moreover, to further diversify the low-level appearances of the collected images, we design a custom variant of \textit{imagecorruptions}~\cite{imagecorruptions} to randomly corrupt 1,012 originally-pristine images from COCO~\cite{cococaps} dataset with one in \textit{15} artificial distortions. The assembled sub-sampled dataset consists of \textbf{18,973} images, which are further fed to human subjects to provide \textit{pathway} feedbacks.

\subsection{Task Definition: the \textbf{\emph{pathway}} Feedbacks}
\label{sec:32}
%\paragraph{Why \textbf{\emph{pathway}} feedbacks instead of attribute scores?}
%The \textbf{\emph{Pathway}} feedbacks indicate the human low-level perception pathway, specifically manifested by initially describing the image's low-level attributes using natural language， and eventually reasoned into corresponding quality assessment.
For the \textbf{Q-Pathway}, to collect a richer and more nuanced understanding of human perception on low-level visual aspects, instead of collecting multi-dimensional scores as in existing studies~\cite{spaq,wu2023explainable}, we opt to collect a new format of annotation, termed \textit{pathway} feedbacks, with an exhaustive natural language description on low-level visual attributes \textit{e.g.~noise, brightness, clarity}) followed by a general conclusion. The rationales for this format are as follows: \textbf{(1)} Primarily, the descriptions can preserve what humans perceive more \textit{completely} and \textit{precisely}. For instance, if an image has both dark and bright areas such as Fig~\ref{fig:three}(a) \textit{upper}, the \underline{brightness} score might not properly record~\cite{paq2piq,wu2022fastervqa} this situation: the positional context cannot be preserved, and the reliability of the score could also be compromised, as neither labeling it as `dark' nor as `bright' is accurate. \textbf{(2)} Moreover, unlike free-form text feedbacks, the order of the two parts in \textit{pathway} feedbacks generally aligns with the human reasoning process. For instance, while human subjects are shown with an \textcolor{red}{\textit{underexposed}} yet \textbf{clear} image, they can provide intuitive reasoning leading to eclectic conclusions like ``\textit{Thus, the quality of the image is acceptable}". This reasoning will help MLLMs to better emulate human perception and understanding related to low-level vision. While this \textit{pathway}-style format faces challenges to be transformed into machine learning objectives in the past, the emergence of MLLMs has provided the opportunity to learn from these direct human feedbacks, in order to allow machines to more precisely and robustly align with human perception.
% context, reasoning, improvement

\subsection{The subjective study process.} 
\label{sec:33}

The subjective study is carried out in a well-controlled laboratory environment, during which a total of 39 \textbf{trained} human subjects are invited. Based on task definition, training material includes not only calibration on \textit{overall quality}, but also on the \textit{respective text descriptions} of different low-level appearances shown in visuals. Furthermore, as the majority of images come from IQA datasets, the mean opinion scores (MOSs) of them are also displayed to subjects to better calibrate them with a common understanding of \textit{quality}. To facilitate their feedback process, we also show a reference attribute set that can be used in the descriptions. To avoid test fatigue of subjects, consecutive feedbacks on more than 30 images will be warned and discouraged; it will be further forcefully paused after 50 images. 58K \textit{pathway} feedbacks are collected during the study, as exemplified in~Fig.~\ref{fig:three}(a).

\begin{figure*}
    \centering
    \includegraphics[width=0.99\linewidth]{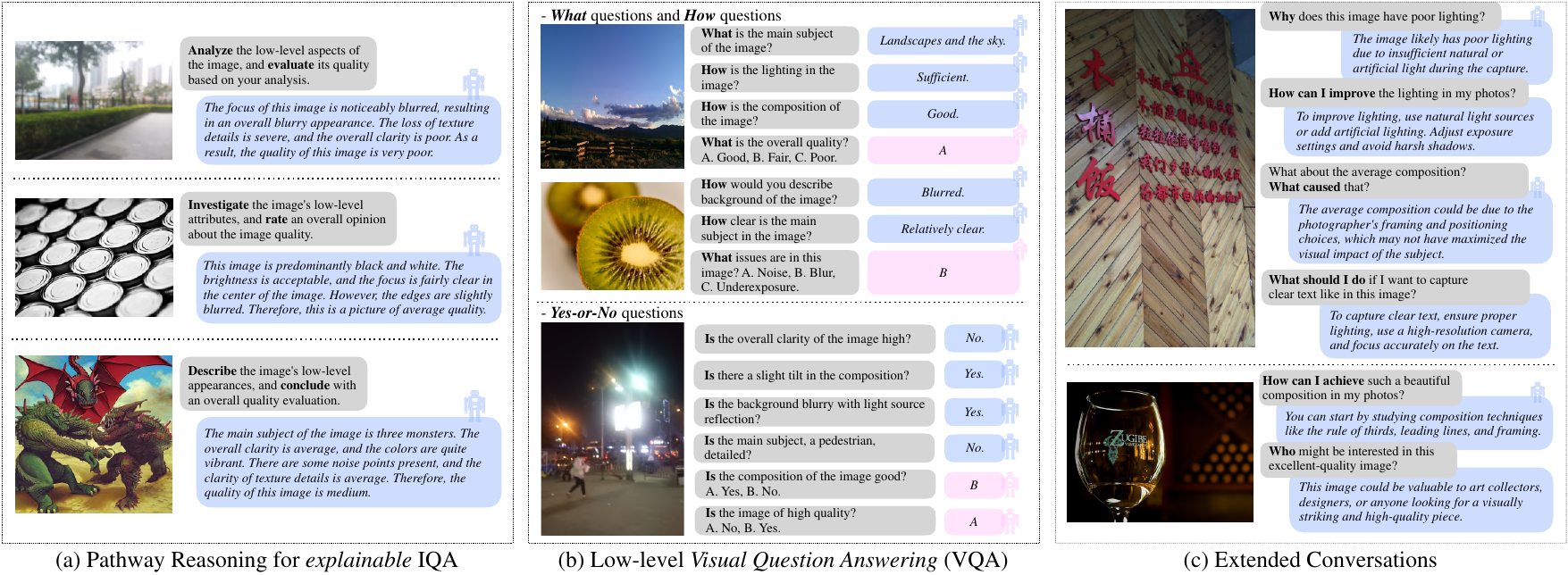}
    \caption{The composition of the \textbf{Q-Instruct} dataset, in which the \textbf{200K} instruction-response pairs include (a) \textbf{58K} pathway reasoning, (b) \textit{visual question answering}, with \textbf{76K} \textit{what/how} questions and \textbf{57K} balanced \textit{yes-or-no} questions, and (c) \textbf{12K} extended conversations. }
    \label{fig:four} 
    \vspace{-10pt}
\end{figure*}

\subsection{Analysis}
\label{sec:34}
% maybe to supp

After the subjective study, we briefly analyze the collected feedbacks. Qualitatively (Fig.~\ref{fig:three}(a)), the \textit{pathway} feedbacks can generally preserve the respective contexts related to low-level attributes. Moreover, feedbacks from different human subjects for the same image (as exemplified in \textsf{[A]}\ and \textsf{[B]} for each image) shows decent consistency (\textit{no controversial information}), and slightly complements one another. Statistically, the length of feedbacks generally ranges from 20 to 100 words, with an average of \textbf{46.4} words, 4 times as long as common high-level image captions~\cite{cococaps,flickrcaps} (Fig~\ref{fig:three}(b)). We also visualize the wordcloud~\cite{1wordcloud} and the bar chart for the top frequency words related to low-level vision\footnote{For better visualization, the two words that appear in every feedback, \textbf{\textit{image}} and \textbf{\textit{quality}}, are removed from the bar chart in Fig.~\ref{fig:three}(d).}, demonstrating that the collected \textbf{Q-Pathway} covers a wide range of low-level attributes, and includes positive and negative feedbacks within similar proportions.

\section{the \textit{Q-Instruct}}

The long and diverse feedbacks in the \textbf{Q-Pathway} provides sufficient reference for the automatic generation process of instruction-response pairs to be used for low-level visual instruction tuning. While the \textit{pathway} feedbacks themselves can teach MLLMs to reason low-level aspects and predict quality (Sec.~\ref{sec:41}), we design more instruction types to allow MLLMs to respond to a variety of human queries, including a \textit{visual question answering} subset (Sec.~\ref{sec:42}) for more accurate low-level perception ability~\cite{wu2023qbench}, and an extended conversation subset (Sec.~\ref{sec:43}) to allow MLLMs to seamlessly \textit{chat} with human about topics related to low-level visual aspects. Overall, the \textbf{Q-Instruct} dataset includes 200K instruction-response pairs, with its details as follows.

%\subsection{Methodology}

\subsection{Low-level Reasoning with \textbf{\textit{pathway}} Feedbacks}
\label{sec:41}

Similar as image captioning~\cite{cococaps,nocaps,flickrcaps}, a general low-level visual description ability is also vital for MLLMs. As analyzed in Fig.~\ref{fig:three}, the pathway \textit{feedbacks} are direct and holistic human responses that generally describe low-level visual appearances. Furthermore, these feedbacks provide \textbf{\textit{reasoning}} from low-level attributes (\textit{brightness, clarity}) to overall quality ratings (\textit{good/poor}), which could activate the potential reasoning abilities~\cite{cot,llmreasoner} of MLLMs on IQA. Henceforth, with each \textit{pathway} feedback as response and a general prompt as instruction, we include \textbf{58K} pathway reasoning (Fig.~\ref{fig:four}(a)) as the primary part of the \textbf{Q-Instruct} dataset.

\subsection{Visual Question Answering (VQA)}
\label{sec:42}

Besides directly apply the \textbf{Q-Pathway} into low-level visual instruction tuning, we also design a GPT~\cite{chatgpt}-participated pipeline to convert them into a \textit{visual question answering} (VQA) subset. In general, we ask GPT to generate diverse-style questions related to low-level-vision from the \textit{pathway} feedbacks, and provide answers with \textit{as few words as possible}. Via this process, we convert the feedbacks into \textbf{76K} questions, including \textit{how} questions answered with opinion-related adjectives (\textit{e.g. good/poor, high/low}), or \textit{i.e.~\textbf{what}} questions answered with attribute-related (\textit{blur/noise/focus}) or context-related (\textit{left/the peacock/the background}) nouns, as shown in the \textit{upper} part of Fig.~\ref{fig:four}(b). We further instruct GPT to generate binary judgments (\textit{yes/no}, Fig.~\ref{fig:four}(b) \textit{lower}) from the feedbacks, and balance \textit{yes} and \textit{no} into 1:1 ratio, with \textbf{57K} \textit{yes-or-no} questions collected at last. As for the answering format, following A-OKVQA~\cite{AOKVQA}, despite the direct answers, we also create several distracting answers for the questions, and convert them into an additional multi-choice question (MCQ) format (\textit{the pink boxes} in Fig.~\ref{fig:four}(b)). %The VQA subset in the \textbf{Q-Instruct} allow MLLMs to more precisely respond to a variety of human queries, expanding their scopes of capabilities and potential applications.

%\paragraph{Building \textit{what/how} questions from simple sentences.} 

%\paragraph{Building \textit{yes-or-no} questions contrastively.}

\subsection{Extended Conversations}
\label{sec:43}

While the first two subsets are designed to enhance the fundamental language-related abilities for low-level vision, the third subset of the \textbf{Q-Instruct}, the \textit{extended conversations} (Fig.~\ref{fig:four}(c)), focuses on improving the ability to discuss with human grounded on the low-level visual aspects of an input image. These discussions include five major scopes: \textbf{1)} Examining the causes of low-level visual patterns; \textbf{2)} Providing improvement suggestions on photography; \textbf{3)} Providing tools to restore, enhance, or edit the image; \textbf{4)} Recommending the image to respective consumers; \textbf{5)} Other conversations that may happen given the low-level visual descriptions provided in the \textit{pathway} feedbacks. Similarly, the extended conversation subset is also generated by GPT, with in total \textbf{12K} conversations collected for the \textbf{Q-Instruct}.

\begin{figure}[t]
  \centering
  \includegraphics[width=\linewidth]{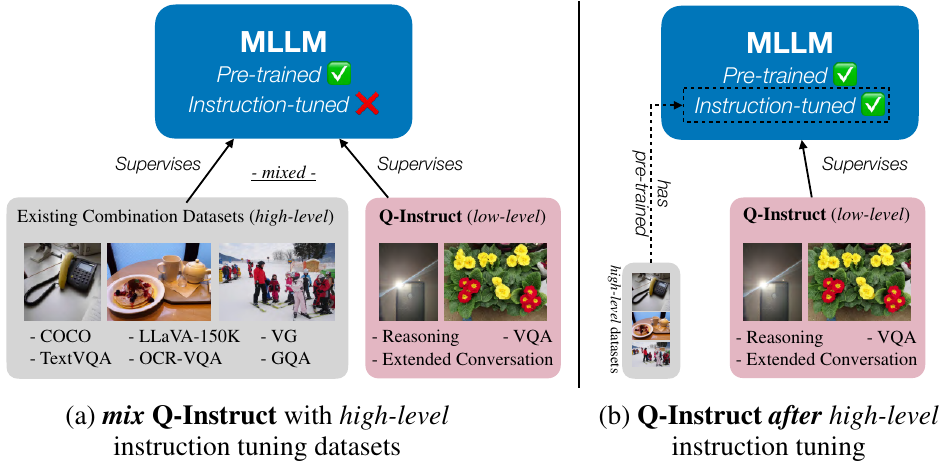}

   \caption{Training strategies for \textit{low-level visual instruction tuning} evaluated in our study, including (a) \textbf{\underline{\textit{mix}}} the \textbf{Q-Instruct} with high-level visual instruction tuning datasets, (b) conduct a further low-level tuning stage with only \textbf{Q-Instruct} \textbf{\underline{\textit{after}}} high-level tuning. }
   \label{fig:five}
   \vspace{-10pt}
\end{figure}

\section{Low-level Visual Instruction Tuning}
\label{sec:5}
In this section, we discuss the standard training strategies for \textit{low-level visual instruction tuning}, \textit{i.e.} when to involve the \textbf{Q-Instruct} dataset during the training of MLLMs. In general, the training of open-source MLLMs~\cite{iblip,otter,xcomposer} includes two stages: \textbf{First,} aligning the representation space of the visual backbone and the LLM with million-scale web data~\cite{laion400m, cc3m}. \textbf{Second,} visual instruction tuning with a combination of human-labeled datasets~\cite{refcoco,cocovqa,cococaps,okvqa}. Considering the scale of the \textbf{Q-Instruct}, 
a general strategy is to \underline{\textbf{\textit{mix}}} its instruction-response pairs with the high-level datasets in the \textbf{second} stage, so as to ideally built their low-level visual abilities within general high-level awareness, as shown in Fig.~\ref{fig:five}(a). Another faster and more convenient strategy is a further \textbf{third} stage only with the \textbf{Q-Instruct} (Fig.~\ref{fig:five}(b)) \underline{\textbf{\textit{after}}} original high-level tuning. In our experiments, we validate that they both bring notable improvements on various low-level visual tasks, and involving \textit{high-level} awareness contributes to the effectiveness of both strategies.

\section{Experiments}

\subsection{Experimental Setups}
\paragraph{Baseline models.} We pick four variants of three state-of-the-art MLLMs within diverse meta structures (Tab.~\ref{tab:arc}) as baseline models to evaluate their low-level visual abilities \textit{before} and \textit{after} training with the \textbf{Q-Instruct}. Each model is evaluated under both strategies as in Fig.~\ref{fig:five}, with the original combination of \textit{high-level} datasets unchanged.
\begin{table}\small
    \centering
    \renewcommand\arraystretch{1.25}
    \renewcommand\tabcolsep{4pt}
    \caption{Baseline MLLMs for \textit{low-level visual instruction tuning}.}
    \vspace{-8pt}
    \resizebox{\linewidth}{!}{\begin{tabular}{l|c|c|c}
\toprule

$^\text{Month/Year}$\textbf{Model Name}& \textbf{Visual Backbone} &  \textbf{V$\to$L Module} & \textbf{Language Model} \\
\hdashline

$^\text{Oct/23}$LLaVA-v1.5 (\textit{7B})~\citep{improvedllava} & CLIP-ViT-L14$^{\uparrow336}$ & MLP & Vicuna-v1.5-7B~\cite{vicuna} \\
$^\text{Oct/23}$LLaVA-v1.5 (\textit{13B})~\citep{improvedllava} & CLIP-ViT-L14$^{\uparrow336}$ & MLP & Vicuna-v1.5-13B~\cite{vicuna} \\
%$^\text{Oct/23}$LLaMA-Adapter-V2.1~\citep{llamaadapterv2} & CLIP-ViT-L14 & cross attention & LLaMA-7B~\cite{llama} \\
$^\text{Oct/23}$mPLUG-Owl-2~\citep{mplugowl2} & CLIP-ViT-L14$^{\uparrow448}$ & Abstractor & LLaMA2-7B~\cite{llama2} \\
$^\text{Sep/23}$InternLM-XComposer-VL~\citep{xcomposer} & EVA-CLIP-G & Perceive Sampler & InternLM-7B~\cite{internlm} \\
%\midrule
%\textcolor{gray}{$^\text{Jan/22}$BLIP~\citep{li2022blip}} & \textcolor{gray}{ViT-B16$^{\uparrow384}$} & \textcolor{gray}{Align\&Fusion} & \textcolor{gray}{BERT-Base-110M~\cite{bert}} \\
%\textcolor{gray}{$^\text{May/22}$CoCa~\citep{coca}} & \textcolor{gray}{ViT-L14} & \textcolor{gray}{Cross Attention} & \textcolor{gray}{Transformer~\cite{allyouneed}} \\
\bottomrule
\end{tabular}}
\label{tab:arc}
\vspace{-10pt}
\end{table}

\begin{table*}\small
    \centering
    \renewcommand\arraystretch{1.1}
    \caption{Comparison of the low-level \textbf{Perception} ability between baseline MLLMs and \textbf{Q-Instruct}-\textit{tuned} versions, on \textbf{LLVisionQA}-{\textit{dev}}. }
    \vspace{-8pt}
    \resizebox{\linewidth}{!}{\begin{tabular}{l|l|ccc|cc|cc|c}
    \toprule
        %\textbf{Sub-categories} & \multicolumn{3}{c|}{\textbf{Question Types}} & \multicolumn{4}{c|}{\textbf{Quadrants of Low-level Concerns}} & \multirow{3}{*}{{\textit{Overall$\uparrow$}}} \\ \cdashline{1-8}
        \textbf{Model} \textit{(variant)} & \textit{Q-Instruct} Strategy  & \multirow{1}{*}{\textit{Yes-or-No$\uparrow$}}& \multirow{1}{*}{\textit{What$\uparrow$}} & \multirow{1}{*}{\textit{How$\uparrow$}} & \multirow{1}{*}{\textit{Distortion$\uparrow$}} & \multirow{1}{*}{\textit{Other$\uparrow$}} & \textit{I-C Distortion$\uparrow$}  &\textit{I-C Other$\uparrow$} & \multirow{1}{*}{{\textit{Overall$\uparrow$}}} \\ \hline
        \textit{random guess} & -- & 50.00\% & 27.86\% & 33.31\% & 37.89\% & 38.48\% & 38.28\% & 35.82\% & 37.80\% \\ \cdashline{1-10}
    \multirow{3}{80pt}{LLaVA-v1.5 (\textit{7B})} & \textit{no} (Baseline) & 66.36\% & 58.19\% & 50.51\% & 49.42\% & {65.74}\% & 54.61\% & {70.61}\% & 58.66\% \\
         & (a) \textbf{\textit{mix}} with high-level & 76.18\%$_\text{\textcolor{red}{+9.82\%}}$&66.37\%$_\text{\textcolor{red}{+8.18\%}}$&57.61\%$_\text{\textcolor{red}{+7.10\%}}$&65.18\%$_\text{\textcolor{red}{+15.76\%}}$&67.59\%$_\text{\textcolor{red}{+1.85\%}}$& 64.80\%$_\text{\textcolor{red}{+10.19\%}}$& 73.06\%$_\text{\textcolor{red}{+2.55\%}}$& 67.09\%$_\text{\textcolor{red}{+8.43\%}}$ \\
     & (b) \textbf{\textit{after}} high-level & 76.91\%$_\text{\textcolor{red}{+10.45\%}}$ & 65.04\%$_\text{\textcolor{red}{+6.85\%}}$ & 55.78\%$_\text{\textcolor{red}{+5.27\%}}$ & 64.01\%$_\text{\textcolor{red}{+14.59\%}}$ & 67.13\%$_\text{\textcolor{red}{+1.39\%}}$ & 64.80\%$_\text{\textcolor{red}{+10.19\%}}$ & 71.84\%$_\text{\textcolor{red}{+1.23\%}}$ & 66.35\%$_\text{\textcolor{red}{+7.69\%}}$ \\ 
 \hdashline
    \multirow{3}{80pt}{LLaVA-v1.5 (\textit{13B})} & \textit{no} (Baseline) & 65.27\% & {64.38}\% & {56.59}\% & 56.03\% & {67.13}\% & {61.18}\% & 67.35\% & {62.14}\% \\
    %[0.76181818 0.65707965 0.59229209]
%[0.64396887 0.69907407 0.625      0.75510204]
%0.6742474916387959
    & (a) \textbf{\textit{mix}} with high-level & 76.18\%$_\text{\textcolor{red}{+10.91\%}}$ & 65.71\%$_\text{\textcolor{red}{+1.33\%}}$ & 59.23\%$_\text{\textcolor{red}{+2.64\%}}$& 64.39\%$_\text{\textcolor{red}{+8.36\%}}$ & 69.91\%$_\text{\textcolor{red}{+2.78\%}}$ & 62.50\%$_\text{\textcolor{red}{+1.32\%}}$ & 75.51\%$_\text{\textcolor{red}{+8.16\%}}$ & 67.42\%$_\text{\textcolor{red}{+5.28\%}}$  \\
    & (b) \textbf{\textit{after}} high-level  & 
76.36\%$_\text{\textcolor{red}{+11.09\%}}$ & 65.04\%$_\text{\textcolor{red}{+0.66\%}}$ & 58.42\%$_\text{\textcolor{red}{+1.83\%}}$ & 65.56\%$_\text{\textcolor{red}{+9.53\%}}$ & 66.44\%$_\text{\textcolor{gray}{-0.69\%}}$  & 64.47\%$_\text{\textcolor{red}{+3.29\%}}$ & 74.29\%$_\text{\textcolor{red}{+6.94\%}}$ &  67.02\%$_\text{\textcolor{red}{+4.88\%}}$  \\ \hdashline
    \multirow{3}{80pt}{mPLUG-Owl-2} & \textit{no} (Baseline) & 72.18\% & 57.96\% & 56.19\% & 56.68\% & 69.21\%& 53.29\%& 72.65\%& 61.61\%  \\
    & (a) \textbf{\textit{mix}} with high-level & 
    75.64\%$_\text{\textcolor{red}{+3.46\%}}$ & 
    67.04\%$_\text{\textcolor{red}{+9.08\%}}$ &  
    59.03\%$_\text{\textcolor{red}{+2.84\%}}$ & 
    71.01\%$_\text{\textcolor{red}{+14.33\%}}$ & 
    65.28\%$_\text{\textcolor{gray}{-3.93\%}}$ &  
    63.16\%$_\text{\textcolor{red}{+9.87\%}}$ & 
    69.80\%$_\text{\textcolor{gray}{-2.85\%}}$ & 
    67.56\%$_\text{\textcolor{red}{+5.95\%}}$ \\
    & (b) \textbf{\textit{after}} high-level & 
    76.00\%$_\text{\textcolor{red}{+3.82\%}}$ & 
    65.04\%$_\text{\textcolor{red}{+7.08\%}}$ & 
    61.66\%$_\text{\textcolor{red}{+5.47\%}}$ &
    65.95\%$_\text{\textcolor{red}{+9.27\%}}$ & 
    68.75\%$_\text{\textcolor{gray}{-0.46\%}}$ &   
    65.46\%$_\text{\textcolor{red}{+12.17\%}}$ & 
    73.88\%$_\text{\textcolor{red}{+1.23\%}}$ & 
    67.96\%$_\text{\textcolor{red}{+6.35\%}}$ \\ \hdashline
    \multirow{3}{80pt}{InternLM-XComposer-VL} & \textit{no} (Baseline) & {69.45}\%  & {65.27}\% & {60.85}\% & {61.67}\% & {70.14}\% & 56.91\% & {75.10}\% & {65.35}\% \\
        & (a) \textbf{\textit{mix}} with high-level & 76.73\%$_\text{\textcolor{red}{+7.28\%}}$ & 69.91\%$_\text{\textcolor{red}{+4.64\%}}$ & 63.89\%$_\text{\textcolor{red}{+3.04\%}}$ & 70.23\%$_\text{\textcolor{red}{+8.56\%}}$ & 71.53\%$_\text{\textcolor{red}{+1.39\%}}$ & 67.43\%$_\text{\textcolor{red}{+10.52\%}}$ & 72.65\%$_\text{\textcolor{gray}{-2.45\%}}$ & 70.43\%$_\text{\textcolor{red}{+5.08\%}}$ \\
    & (b) \textbf{\textit{after}} high-level &  78.36\%$_\text{\textcolor{red}{+8.91\%}}$ & 68.58\%$_\text{\textcolor{red}{+3.31\%}}$ & 63.08\%$_\text{\textcolor{red}{+2.23\%}}$ & 
65.37\%$_\text{\textcolor{red}{+3.70\%}}$ &  73.15\%$_\text{\textcolor{red}{+3.01\%}}$ & 68.42\%$_\text{\textcolor{red}{+11.51\%}}$ & 78.37\%$_\text{\textcolor{red}{+3.27\%}}$ & 70.37\%$_\text{\textcolor{red}{+5.02\%}}$\\ 
        \bottomrule
    \end{tabular}}
    \vspace{-6pt}
    \label{tab:perception}
\end{table*}

\begin{table*}\small
    \centering
    \renewcommand\arraystretch{1.1}
    \caption{Comparison of the low-level \textbf{Perception} ability between baseline MLLMs and \textbf{Q-Instruct}-\textit{tuned} versions, on \textbf{LLVisionQA}-{\textit{test}}. }
    \vspace{-8pt}
    \resizebox{\linewidth}{!}{\begin{tabular}{l|l|ccc|cc|cc|c}
    \toprule
        %\textbf{Sub-categories} & \multicolumn{3}{c|}{\textbf{Question Types}} & \multicolumn{4}{c|}{\textbf{Quadrants of Low-level Concerns}} & \multirow{3}{*}{{\textit{Overall$\uparrow$}}} \\ \cdashline{1-8}
        \textbf{Model} \textit{(variant)} & \textit{Q-Instruct} Strategy  & \multirow{1}{*}{\textit{Yes-or-No$\uparrow$}}& \multirow{1}{*}{\textit{What$\uparrow$}} & \multirow{1}{*}{\textit{How$\uparrow$}} & \multirow{1}{*}{\textit{Distortion$\uparrow$}} & \multirow{1}{*}{\textit{Other$\uparrow$}} & \textit{I-C Distortion$\uparrow$}  &\textit{I-C Other$\uparrow$} & \multirow{1}{*}{{\textit{Overall$\uparrow$}}} \\ \hline
        \textit{random guess} & -- & 50.00\% & 28.48\% & 33.30\% & 37.24\% & 38.50\% & 39.13\% & 37.10\% & 37.94\% \\ \cdashline{1-10}
        \multirow{3}{80pt}{LLaVA-v1.5 (\textit{7B})} & \textit{no} (Baseline) & 64.60\% & 59.22\% & 55.76\% & 47.98\% & {67.30}\% & {58.90}\% & {73.76}\% & 60.07\% \\
            & (a) \textbf{\textit{mix}} with high-level & 78.65\%$_\text{\textcolor{red}{+14.05\%}}$ & 63.99\%$_\text{\textcolor{red}{+4.77\%}}$ & 63.79\%$_\text{\textcolor{red}{+8.03\%}}$ & 65.26\%$_\text{\textcolor{red}{+17.28\%}}$ & 68.97\%$_\text{\textcolor{red}{+1.67\%}}$ & 67.81\%$_\text{\textcolor{red}{+8.91\%}}$ & 79.47\%$_\text{\textcolor{red}{+5.71\%}}$ & 69.30\%$_\text{\textcolor{red}{+9.23\%}}$ \\
     & (b) \textbf{\textit{after}} high-level & 78.46\%$_\text{\textcolor{red}{+13.86\%}}$& 63.34\%$_\text{\textcolor{red}{+4.12\%}}$& 58.85\%$_\text{\textcolor{red}{+3.09\%}}$& 60.46\%$_\text{\textcolor{red}{+12.48\%}}$& 68.74\%$_\text{\textcolor{red}{+1.44\%}}$ & 69.52\%$_\text{\textcolor{red}{+10.62\%}}$ & 76.81\%$_\text{\textcolor{red}{+3.05\%}}$ & 67.42\%$_\text{\textcolor{red}{+7.35\%}}$ \\
  \hdashline
        \multirow{3}{80pt}{LLaVA-v1.5 (\textit{13B})} & \textit{no} (baseline) & 64.96\% & {64.86}\% & 54.12\% & 53.55\% & {66.59}\% & {58.90}\% & 71.48\% & 61.40\% \\
        %[0.77189781 0.68546638 0.65432099]
%[0.64683301 0.71121718 0.67465753 0.85551331]
%0.7070234113712375
    & (a) \textbf{\textit{mix}} with high-level & 77.19\%$_\text{\textcolor{red}{+13.23\%}}$ & 68.55\%$_\text{\textcolor{red}{+3.69\%}}$ & 65.43\%$_\text{\textcolor{red}{+11.31\%}}$ & 64.68\%$_\text{\textcolor{red}{+11.13\%}}$ & 71.12\%$_\text{\textcolor{red}{+4.43\%}}$ & 67.47\%$_\text{\textcolor{red}{+8.57\%}}$ & 85.55\%$_\text{\textcolor{red}{+14.07\%}}$ & 70.70\%$_\text{\textcolor{red}{+9.30\%}}$  \\
    & (b) \textbf{\textit{after}} high-level & 80.66\%$_\text{\textcolor{red}{+15.70\%}}$ & 67.25\%$_\text{\textcolor{red}{+2.39\%}}$ & 61.93\%$_\text{\textcolor{red}{+7.81\%}}$ & 66.03\%$_\text{\textcolor{red}{+12.48\%}}$ & 70.41\%$_\text{\textcolor{red}{+3.82\%}}$ & 69.86\%$_\text{\textcolor{red}{+10.96\%}}$ & 79.85\%$_\text{\textcolor{red}{+8.37\%}}$ & 70.43\%$_\text{\textcolor{red}{+9.03\%}}$\\ \hdashline
    \multirow{3}{80pt}{mPLUG-Owl-2} & \textit{no} (Baseline) &  72.26\% & 55.53\% & 58.64\% & 52.59\% & 71.36\% & 58.90\% & 73.00\% & 62.68\% \\
    & (a) \textbf{\textit{mix}} with high-level & 
    78.47\%$_\text{\textcolor{red}{+6.21\%}}$&  
    67.90\%$_\text{\textcolor{red}{+12.37\%}}$ & 
    63.37\%$_\text{\textcolor{red}{+4.73\%}}$ &
    68.52\%$_\text{\textcolor{red}{+15.93\%}}$ & 
    68.02\%$_\text{\textcolor{gray}{-3.34\%}}$ &  
    70.21\%$_\text{\textcolor{red}{+11.31\%}}$ & 
    77.57\%$_\text{\textcolor{red}{+4.57\%}}$ & 
    70.30\%$_\text{\textcolor{red}{+7.62\%}}$  \\
    & (b) \textbf{\textit{after}} high-level &  
    78.47\%$_\text{\textcolor{red}{+6.21\%}}$ & 
    60.74\%$_\text{\textcolor{red}{+5.21\%}}$ &  
    66.46\%$_\text{\textcolor{red}{+7.82\%}}$ &
    63.34\%$_\text{\textcolor{red}{+10.75\%}}$ &  
    71.36\%$_\text{\textcolor{red}{$\pm0$}}$ & 
    68.15\%$_\text{\textcolor{red}{+9.25\%}}$ & 
    77.95\%$_\text{\textcolor{red}{+4.95\%}}$ & 
    69.10\%$_\text{\textcolor{red}{+6.42\%}}$ \\ \hdashline
        \multirow{3}{80pt}{InternLM-XComposer-VL} & \textit{no} (Baseline) & 68.43\% & {62.04}\% & {61.93}\% & {56.81}\% & {70.41}\% & 57.53\% & {77.19}\% & {64.35}\% \\
            & (a) \textbf{\textit{mix}} with high-level & 78.65\%$_\text{\textcolor{red}{+10.22\%}}$ & 68.33\%$_\text{\textcolor{red}{+6.29\%}}$ &  66.26\%$_\text{\textcolor{red}{+4.33\%}}$ & 70.24\%$_\text{\textcolor{red}{+13.43\%}}$ &  71.12\%$_\text{\textcolor{red}{+0.81\%}}$ &  68.15\%$_\text{\textcolor{red}{+10.62\%}}$ & 77.95\%$_\text{\textcolor{red}{+0.76\%}}$ & 71.44\%$_\text{\textcolor{red}{+7.09\%}}$ \\
    & (b) \textbf{\textit{after}} high-level & 79.56\%$_\text{\textcolor{red}{+11.13\%}}$ & 64.64\%$_\text{\textcolor{red}{+2.60\%}}$ &  65.43\%$_\text{\textcolor{red}{+3.50\%}}$ & 
64.30\%$_\text{\textcolor{red}{+7.49\%}}$ & 71.60\%$_\text{\textcolor{red}{+1.19\%}}$ & 66.44\%$_\text{\textcolor{red}{+8.91\%}}$ & 84.79\%$_\text{\textcolor{red}{+7.60\%}}$ & 70.37\%$_\text{\textcolor{red}{+6.02\%}}$ \\ 
       \bottomrule
    \end{tabular}}
    \vspace{-6pt}
    \label{tab:perceptiontest}
\end{table*}

%In addition to powerful MLLMs, we also investigate how the smaller-scale image-to-text model, \textbf{{BLIP}}~\cite{li2022blip} (2022), performs while fine-tuned with the \textbf{Q-Pathway} by considering the \textit{pathway} feedbacks as a special type of image captions, which could bring multi-modality low-level description ability to scenarios with tighter computational limits.%or the \textit{visual question answering} subset of the \textbf{Q-Instruct}. 

\paragraph{Training Settings.} We follow the default instruction tuning hyper-parameters of MLLMs during all training processes involving the \textbf{Q-Instruct}. As we aim to reach a unified low-level visual foundation model, for each MLLM, the final checkpoint is saved and tested for all evaluations. To avoid data contamination, during training, we remove data items with images that may appear in the evaluation sets.%For LLaVA-v1.5, as we notice a narrow performance gap on low-level tasks between the two variants, we only fine-tune the lightweight \textit{7B} version with the \textbf{Q-Instruct}.

 %Moreover, we \textit{qualitatively} study whether and how fine-tuned MLLMs can converse with human related to low-level visual aspects. Despite the specific low-level abilities, we also briefly measure whether the fine-tuned MLLMs can preserve general vision-language abilities through MMBench~\cite{mmbench}, a holistic benchmark designed for general multi-modality scenarios. 

\subsection{Main Results}

The low-level visual abilities of MLLMs after \textit{low-level visual instruction tuning} are quantitatively evaluated in three tasks defined by~\cite{wu2023qbench}, including \textbf{(A1) Perception}, by measuring the accuracy of answering multi-choice questions (MCQ) related to low-level vision (Fig.~\ref{fig:one}); \textbf{(A2) Description}, which examines how MLLMs can generally transform low-level visual information into text. As for \textbf{(A3) Quality Assessment}, considering that the \textbf{Q-Instruct} already contains a large proportion of images in major IQA databases, we evaluate and discuss how the instructed MLLMs generalize on unseen images. For reproducibility, all responses from MLLMs are generated with \textit{greedy search}. Qualitative analyses are provided in supplementary materials.

\paragraph{(A1) Perception (MCQ).} From Tab.~\ref{tab:perception} and Tab.~\ref{tab:perceptiontest}, we observe that either strategy of including \textbf{Q-Instruct} into the training of MLLMs can significantly improve their low-level perception ability. The results demonstrate the effectiveness of the proposed pipeline to automatically generate the VQA subset (\textit{including MCQ}) from the pathway feedbacks via GPT, which could be expected to extend to further query types. Specifically, among all dimensions, we notice that the accuracy on \textit{Yes-or-No} question type is most significantly enhanced (\textit{avg. more than 10\%}). Moreover, improvements on \textbf{distortions} are more significant than on \textbf{other} low-level attributes (\textit{aesthetics, photography techniques}), suggesting that the major concerns as raised by human in the \textbf{Q-Pathway} are still related to distortions. We hope that our pipeline can be extended to cover more types of questions and a broader range of concerns in the future.

\paragraph{(A2) Description.}The \textit{low-level visual instruction tuning} also notably improve the low-level description ability of MLLMs, especially on the \textit{relevance} (\textit{+0.31}), with all \textit{tuned} variants obtaining more than 1.5/2 average score.  In contrast, the improvements on \textit{completeness} (\textit{+0.17}) and \textit{precision} (\textit{+0.04}) are less significant, implying that the \textbf{captioning-like} instruction format may not be sufficient for the low-level description task that requires \textit{much longer} responses. We look forward to better solutions in the future.
\begin{table}\small
    \centering
    \renewcommand\arraystretch{1.16}
        \renewcommand\tabcolsep{4pt}
        \caption{Comparison of the low-level \textbf{Description} ability between baseline MLLMs and \textbf{Q-Instruct}-\textit{tuned} versions, under the same prompt: \textit{``Describe and evaluate the quality of the image."}}
        \vspace{-8pt}
    \resizebox{\linewidth}{!}{\begin{tabular}{l|l|c|c|c|c}
    \toprule
        \textbf{Model} \textit{(variant)} & \textit{Q-Instruct} Strategy & \multicolumn{1}{c|}{{\textit{completeness}}} & \multicolumn{1}{c|}{{\textit{precision}}} & \multicolumn{1}{c|}{{\textit{relevance}}} & \textit{\textbf{sum}} \\ \hline
        \multirow{3}{70pt}{LLaVA-v1.5 (\textit{7B})} & \textit{no} (Baseline) &  0.90 & 1.13 & 1.18 & 3.21  \\
        & (a) \textbf{\textit{mix}} w/ high-level & \textbf{1.12} & \textbf{1.17}  & \textbf{1.57} & \textbf{3.86}\\
        & (b) \textbf{\textit{after}} high-level & 1.11 & 1.16 & 1.54 & 3.82\\ \hdashline
        \multirow{3}{70pt}{LLaVA-v1.5 (\textit{13B})} & \textit{no} (Baseline) &   0.91 & {1.28}  & 1.29 & 3.47 \\
        & (a) \textbf{\textit{mix}} w/ high-level & \textbf{1.14} & \textbf{1.29} & 1.58 & \textbf{4.01} \\
        & (b) \textbf{\textit{after}} high-level & 1.13 & 1.26 & \textbf{1.61} & 4.00 \\ \hdashline
        \multirow{3}{70pt}{mPLUG-Owl-2} & \textit{no} (Baseline) & 1.06 & 1.24 & 1.36 &  3.67  \\
        & (a) \textbf{\textit{mix}} w/ high-level &  \textbf{1.18} & \textbf{1.29} &  \textbf{1.57} & \textbf{4.04} \\
        & (b) \textbf{\textit{after}} high-level & 1.16 & 1.27 & \textbf{1.57} & 3.99 \\ \hdashline
        \multirow{3}{70pt}{InternLM-XComposer-VL} & \textit{no} (Baseline) & 1.03 & 1.26 & 1.27 & 3.56  \\
        & (a) \textbf{\textit{mix}} w/ high-level & {1.16} & \textbf{1.35} & \textbf{1.63} & \textbf{4.14}\\
        & (b) \textbf{\textit{after}} high-level & \textbf{1.18} & 1.34 & 1.62 & \textbf{4.14} \\ \hline
\multicolumn{2}{l|}{\textit{Average Improvement}} & \textit{+0.17} & \textit{+0.04} & \textit{+0.31} 	& \textit{+0.52} 
\\
        \bottomrule
    \end{tabular}}
    \vspace{-15pt} 
    \label{tab:description}
\end{table}
\begin{table*}\small
    \centering
    \renewcommand\arraystretch{1.15}
    \renewcommand\tabcolsep{4pt}
    \caption{Comparison of the \textbf{Quality Assessment (A3)} ability between baseline MLLMs and \textbf{Q-Instruct}-\textit{tuned} versions, where \textit{``Mostly Seen"} datasets denote those with the majority of their images sampled in the Q-Instruct, and \textit{``Barely Seen"} represent those with only a small proportion ($<$20\%) sampled. The \textit{``Never Seen"} datasets have \textbf{zero} overlap with the \textbf{Q-Instruct}. Metrics are SRCC~/~PLCC.}
    \vspace{-5pt}
    \resizebox{\linewidth}{!}{\begin{tabular}{l:l|cc:ccc|ccc}
    \toprule
    \multicolumn{2}{c|}{\textbf{Dataset Group}} &   \multicolumn{2}{c:}{\textit{Mostly Seen}} & \multicolumn{3}{c|}{{\textit{Barely Seen}}} & \multicolumn{3}{c}{{\textit{Never Seen}}}\\ \cdashline{1-10}
    \multicolumn{2}{c|}{\textit{\% of dataset seen during training}} &  48.92\% & 95.26\% &  2.00\% & 17.11\% & 13.41\% & \textbf{0\%} & \textbf{0\%}  & \textbf{0\%} \\  \cdashline{1-10}
     \textbf{Model} (\textit{variant}) & \textit{Q-Instruct} Strategy  &{\textit{KonIQ-10k}} & {\textit{SPAQ}} & {\textit{LIVE-FB}} & \textit{LIVE-itw} & {\textit{AGIQA-3K}} & {\textit{CGIQA-6K}}  & {\textit{KADID-10K}} & {\textit{KonViD-1k}}\\ \hline 
    NIQE & --& 0.316~/~0.377 & {0.693}~/~{0.669} & 0.211~/~0.288 & {0.480}~/~0.451 & 0.562~/~0.517  & 0.075~/~0.056 & 0.374~/~0.428 & 0.541~/~0.553 \\ \hdashline
\multirow{3}{60pt}{LLaVA-v1.5 (\textit{7B)}} & \textit{no} (Baseline) & {0.463}~/~0.459 & 0.443~/~0.467  & {0.310}~/~0.339 & 0.445~/~0.481  & {0.664}~/~{0.754} & {0.285}~/~{0.297} & 0.390~/~0.400 & 0.461~/~0.495  \\
& (a) \textbf{\textit{mix}} w/ high-level & 0.809~/~0.852 & 0.880~/~0.883 & 0.377~/~0.436 &  0.800~/~0.806 & 0.724~/~0.828   &  0.521~/~0.535 & 0.688~/~0.695  & 0.766~/~0.717 \\
& (b) \textbf{\textit{after}} high-level & 0.793~/~0.850 & 0.887~/~0.888 & 0.385~/~0.447  & 0.805~/~0.810  & 0.729~/~0.830 & 0.501~/~0.524 & 0.695~/~0.702 & \textbf{0.780}~/~\textbf{0.731} \\ \hdashline

\multirow{3}{60pt}{LLaVA-v1.5 (\textit{13B)}} & \textit{no} (Baseline) & {0.471}~/~0.541 & 0.563~/~0.584  & {0.305}~/~0.321 & 0.344~/~0.358  & {0.672}~/~{0.738} & {0.321}~/~{0.333} & 0.417~/~0.440 & 0.518~/~0.577\\
& (a) \textbf{\textit{mix}} w/ high-level & 0.732~/~0.787 & 0.858~/~0.848 & 0.371~/~0.463 & 0.629~/~0.701 & 0.709~/~0.814 & 0.471~/~0.488 & 0.627~/~0.626 & 0.720~/~0.733 \\
& (b) \textbf{\textit{after}} high-level & 0.748~/~0.798 & 0.867~/~0.869 & 0.359~/~0.417 & 0.695~/~0.719 & 0.696~/~0.766 & 0.494~/~0.516 & 0.633~/~0.641 & 0.706~/~0.692 \\ \hdashline
\multirow{3}{60pt}{mPLUG-Owl-2} & \textit{no} (Baseline) &  0.196~/~0.252 & 0.589~/~0.614 & 0.217~/~0.286 & 0.293~/~0.342 & 0.473~/~0.492 & -0.024~/~-0.032 &  0.541~/~0.546 & 0.409~/~0.442\\
& (a) \textbf{\textit{mix}} w/ high-level &  0.899~/~0.916 & 0.899~/~\textbf{0.903} & 0.432~/~0.545 & 0.829~/~0.822 & 0.743~/~0.806 & \textbf{0.624}~/~\textbf{0.636} & 0.698~/~0.676 & 0.693~/~0.663\\
& (b) \textbf{\textit{after}} high-level & \textbf{0.911}~/~\textbf{0.921} & 0.901~/~0.898 & {0.442}~/~\textbf{0.535} & \textbf{0.842}~/~\textbf{0.840} & 0.700~/~0.763 & 0.572~/~0.578 & 0.682~/~0.683 & 0.769~/~0.721 \\ \hdashline
\multirow{3}{60pt}{InternLM-XComposer-VL} & \textit{no} (Baseline) & 0.568~/~0.616 & 0.731~/~0.751 & 0.358~/~0.413 & 0.619~/~0.678 & 0.734~/~0.777 & 0.246~/~0.268 & 0.540~/~0.563 & 0.620~/~0.649 \\
& (a) \textbf{\textit{mix}} w/ high-level & 0.874~/~0.892 & \textbf{0.909}~/~0.897 & 0.442~/~0.518 & 0.820~/~0.811 & \textbf{0.785}~/~0.830 & 0.391~/~0.411 & \textbf{0.706}~/~\textbf{0.710} & 0.739~/~0.702 \\
& (b) \textbf{\textit{after}} high-level & 0.816~/~0.858 & 0.879~/~0.884 & \textbf{0.443}~/~0.510 &  0.771~/~0.801 & 0.772~/~\textbf{0.847} & 0.394~/~0.420 &  0.677~/~0.645 &  0.743~/~0.730 \\ \hline
\multicolumn{2}{l|}{\textit{Average Improvement}} & \textit{+0.398/+0.392} & 	\textit{+0.304/+0.280} &	\textit{+0.108/+0.144} &	\textit{+0.349/+0.324} &	\textit{+0.097/+0.120} &	\textit{+0.289/+0.297} &	\textit{+0.204/+0.185} & \textit{+0.238/+0.170} 
\\
    \bottomrule
  \end{tabular}}
    \vspace{-10pt}
    \label{tab:assessment}
\end{table*}
\noindent\paragraph{(A3) Image Quality Assessment (IQA).} Despite the two directly tuned tasks, we follow the \textit{softmax} pooling strategy~\cite{wu2023qbench} to extract quality scores from MLLMs and evaluate their IQA ability, as listed in Tab.~\ref{tab:assessment}. Primarily, we notice the excellent performance on two ``\textit{mostly seen}" datasets. As we do not directly use any MOS values during training, this result suggests that we can effectively tune MLLMs to reach very high accuracy on IQA \textbf{without any numerical values} as supervision. This result by-side suggests the high reliability of the proposed datasets. The more exciting results are the huge improvements on ``\textit{barely seen}'' (with a small proportion of images sampled into the \textbf{Q-Instruct}) and even ``\textit{never seen}" (cross-set) datasets. Considering the three ``\textit{never seen}" datasets~\cite{zhang2023subjective,kv1k,kadid} (\textit{with computer-generated images, artificially-degraded image, and even videos respectively}) have notable domain gap with the major part of the \textbf{Q-Instruct} dataset (\textit{mostly in-the-wild photographs}), the {\textit{+0.243}} average SRCC gain on them demonstrates that the \textit{low-level instruction tuning} can robustly improve low-level perception abilities of MLLMs on a broad domain.

%\subsection{Preservation of General Abilities}

%\subsection{Results of BLIP}

\subsection{Ablation Studies}

\begin{figure}[t]
  \centering
  \includegraphics[width=\linewidth]{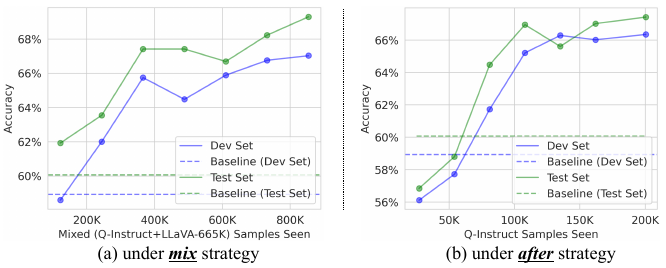}
    \vspace{-15pt}
   \caption{Accuracy on MCQ questions with respect to data samples seen during training (\textit{in comparison with baseline}), demonstrating the effectiveness of scaling up the \textbf{Q-Instruct} dataset.}
   \label{fig:seven}
   \vspace{-12pt}
\end{figure}

Despite the main results for \textit{low-level visual instruction tuning}, we also compare among several data variations during tuning on {LLaVA-v1.5} (\textit{7B}), analyzed as follows.

\paragraph{\#1: Effects of scaling up the Q-Instruct.} The first group of variations discuss the effects of data amount during \textit{low-level visual instruction tuning}. As illustrated in Fig.~\ref{fig:seven}, under either \textbf{\textit{mix}} or \textbf{\textit{after}} strategy, scaling up the \textbf{Q-Instruct} during training can continuously improve the low-level perceptual accuracy. Moreover, the results suggest that the performance of MLLMs is still not saturated even with the current 200K data scale, encouraging us to further unleash their vast underlying power on tackling low-level visual tasks.

\begin{table}\small
    \centering
    \renewcommand\arraystretch{1.13}
        \caption{Comparison on low-level \textbf{Description} ability between \textit{full} \textbf{Q-Instruct} and \textit{only} \textbf{Q-Pathway} as low-level training dataset.}
        \vspace{-8pt}
    \resizebox{\linewidth}{!}{\begin{tabular}{l|l|c|c|c|c}
    \toprule
          \textit{Q-Instruct} Strategy & low-level dataset & \multicolumn{1}{c|}{{\textit{completeness}}} & \multicolumn{1}{c|}{{\textit{precision}}} & \multicolumn{1}{c|}{{\textit{relevance}}} & \textit{\textbf{sum}} \\ \hline
         \textit{no} (Baseline) & None &  0.90 & 1.13 & 1.18 & 3.21  \\ \hdashline
         \multirow{2}{78pt}{(a) \textbf{\textit{mix}} w/ high-level} & \textit{only} \textbf{Q-Pathway} & 1.07 & 1.13 & 1.54 & 3.74\\
         &  \textit{full} \textbf{Q-Instruct} & \textbf{1.12} & \textbf{1.17}  & \textbf{1.57} & \textbf{3.86}\\ \hdashline
         \multirow{2}{78pt}{(b) \textbf{\textit{after}} high-level} & \textit{only} \textbf{Q-Pathway} & 1.02 & 1.12 & \textbf{1.55} & 3.69 \\
        & \textit{full} \textbf{Q-Instruct} & \textbf{1.11} & \textbf{1.16} & 1.54 & \textbf{3.82}\\

        \bottomrule
    \end{tabular}}
    \vspace{-6pt} 
    \label{tab:unifiedvsseparated}
\end{table}

\begin{table}\small
    \centering
    \renewcommand\arraystretch{1.13}
        \caption{Comparison on low-level \textbf{Perception} ability (\textit{test set}) between training with \textit{full} \textbf{Q-Instruct} dataset and \textit{only} VQA subset.}
        \vspace{-8pt}
    \resizebox{\linewidth}{!}{\begin{tabular}{l|l|c|c|c|c}
    \toprule
          \textit{Q-Instruct} Strategy & low-level dataset & \multicolumn{1}{c|}{{\textit{Yes-or-No}}} & \multicolumn{1}{c|}{{\textit{What}}} & \multicolumn{1}{c|}{{\textit{How}}} & \textit{Overall} \\ \hline
         \textit{no} (Baseline) & None & 64.6\% & 59.2\% & 55.8\% & 60.1\%  \\ \hdashline
         \multirow{2}{78pt}{(a) \textbf{\textit{mix}} w/ high-level} & \textit{only} VQA subset  & 78.1\% & 61.5\% & 61.5\% &  67.6\% \\
         &  \textit{full} \textbf{Q-Instruct} & \textbf{78.7\%} & \textbf{64.0\%} & \textbf{63.8\%} & \textbf{69.3\%} \\ \hdashline
         
         \multirow{2}{78pt}{(b) \textbf{\textit{after}} high-level} & \textit{only} VQA subset & 77.9\% & 61.8\% & 56.8\% &   66.1\% \\
        & \textit{full} \textbf{Q-Instruct} & \textbf{78.5\%} & \textbf{63.3\%} & \textbf{58.9\%}  & \textbf{67.4\%} \\

        \bottomrule
    \end{tabular}}
    \vspace{-10pt} 
    \label{tab:unifiedvsseparatedvqa}
\end{table}

\paragraph{\#2: Effects of joint training.} In the \textit{low-level visual instruction tuning}, we combine different subsets together and train them jointly under one unified model. To validate its effectiveness, we compare this approach with traditional task-separate tuning, on both low-level description (Tab.~\ref{tab:unifiedvsseparated}) and question-answering (Tab.~\ref{tab:unifiedvsseparatedvqa}) capabilities. Both experiments indicate that a joint learning scheme can improve the accuracy on these abilities, especially when low-level data is independently used during tuning. While the different subsets in the \textbf{Q-Instruct} come from the same original human feedbacks, the improvement is cost-efficient, and inspires further explorations for \textit{low-level visual instruction tuning} to expand to even more tasks, so as to further improve the low-level capabilities of these MLLMs.

\begin{table}\small
    \centering
    \renewcommand\arraystretch{1.13}
        \caption{Comparison between the proposed two strategies (as in Sec.~\ref{sec:5}, and another variant that \textbf{\textit{replaces}} high-level tuning into the low-level tuning, on their low-level \textbf{Perception} ability (\textit{test set}).}
        \vspace{-5pt}
    \resizebox{\linewidth}{!}{\begin{tabular}{l|c|c|c|c}
    \toprule
          \textit{Q-Instruct} Strategy  & \multicolumn{1}{c|}{{\textit{Yes-or-No}}} & \multicolumn{1}{c|}{{\textit{What}}} & \multicolumn{1}{c|}{{\textit{How}}} & \textit{Overall} \\ \hline
         \textit{no} (Baseline)  & 64.6\% & 59.2\% & 55.8\% & 60.1\%  \\ \hdashline
         {\textbf{\textit{replace}} high-level} (\textit{not adopted}) & 75.0\% & 59.4\% & 56.4\% & 64.1\% \\ \hdashline
         \textbf{\textit{mix}} with high-level (\textit{ours}, strategy (a)) & \textbf{78.7\%} & \textbf{64.0\%} & \textbf{63.8\%} & \textbf{69.3\%} \\
         \textbf{\textit{after}} high-level (\textit{ours}, strategy (b)) &  \textbf{78.5\%} & \textbf{63.3\%} & \textbf{58.9\%}  & \textbf{67.4\%} \\ 
        \bottomrule
    \end{tabular}}
    \vspace{-8pt} 
    \label{tab:nohighlevel}
\end{table}
\paragraph{\#3: Effects of high-level awareness.} While we notice generally on par abilities between the \textbf{\textit{mix}} strategy and the \textbf{\textit{after}} strategy, we further investigate the performance if we \textbf{\textit{replace}} the second stage datasets into the \textbf{Q-Instruct}, while no high-level instruction tuning datasets are involved during training. As compared in Tab.~\ref{tab:nohighlevel}, the ``\textit{replace}'' strategy is notably worse than the two adopted strategies in Sec.~\ref{sec:5}, suggesting that fundamental high-level awareness is important on general low-level visual recognition for MLLMs.

%\subsection{Qualitative Analysis}

\section{Conclusion}
Our work proposes the first-of-a-kind multi-modal datasets on low-level visual aspects, including the \textbf{Q-Pathway} with \textbf{58K} human \textit{text} feedbacks, and the derived \textbf{Q-Instruct} with \textbf{200K} instruction-response pairs, to facilitate \textit{low-level visual instruction tuning} for MLLMs. They allow MLLMs to significantly improve their question-answering accuracy related to low-level visual perception, and showcase the potential for providing more reliable low-level descriptions for images and eventually relieving human burdens on this task.
Further, their IQA performance reveals an intriguing phenomenon, that pure \textit{text-driven} instruction tuning can sufficiently align MLLMs with numerical quality scores, with impressive generalization on unseen types of visual inputs. In summary, our work has advanced a solid step forward on improving the low-level visual abilities of MLLMs, and we hope that our progress and insights can encourage future explorations towards an eventual goal that foundation models understand the low-level visual world like a human.

\newpage

{
    \small
    \bibliographystyle{ieeenat_fullname}
    \bibliography{main}
}

% WARNING: do not forget to delete the supplementary pages from your submission 
% \input{sec/X_suppl}

\clearpage
\maketitlesupplementary
\appendix

\section{Details for Data Collection}
\label{sec:suppdata}

\subsection{Interface for Subjective Experiments}

\begin{figure*}
    \includegraphics[width=\textwidth]{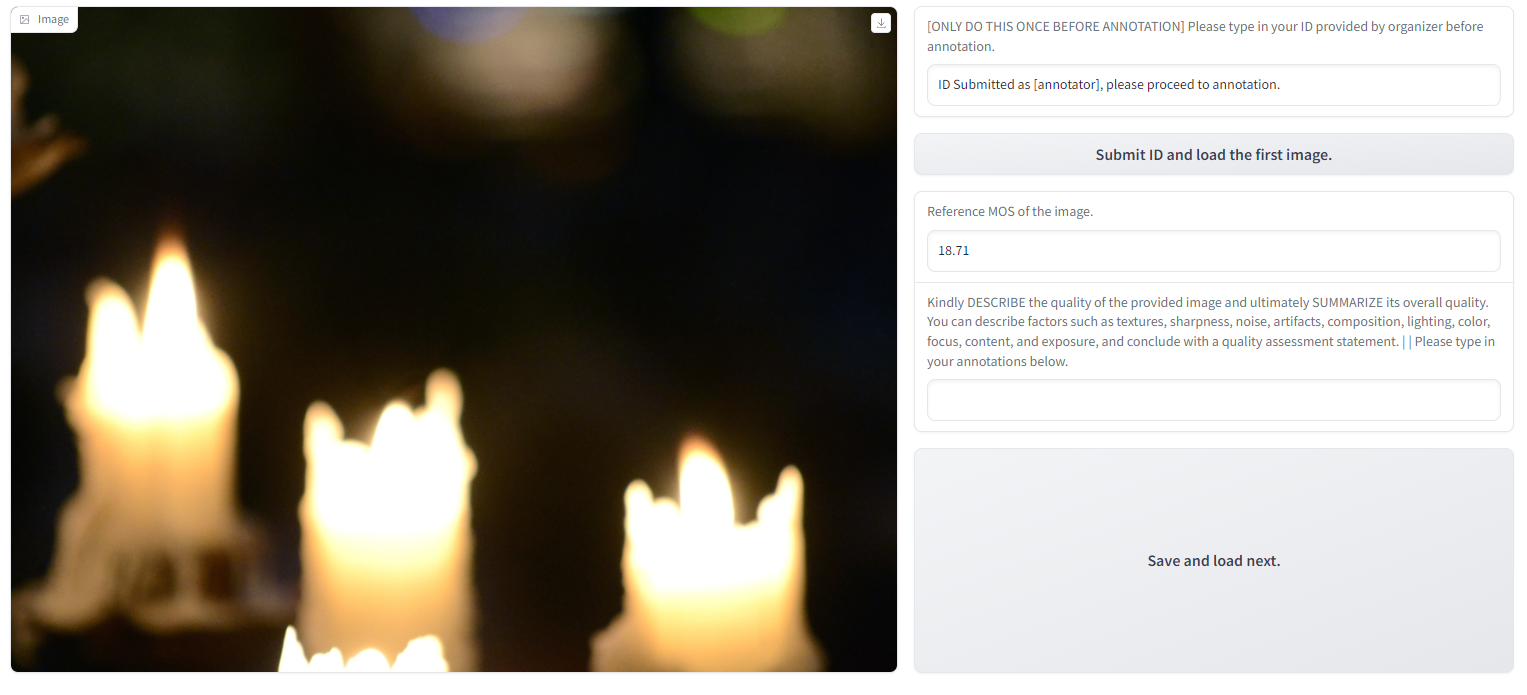}
    \vspace{-16pt}
    \caption{The gradio interface for subjects to provide \textit{\textbf{pathway}} feedbacks. While the quality scores (MOS) of images are available, these scores will be provided to the subjects as a reference, allowing the feedbacks to truly become explanations of these quality scores. }
    \label{fig:interface}
    \vspace{-10pt}
\end{figure*}

The interface for the subjective experiments is built upon Gradio 3.34.0, set up locally on Ubuntu 20.04 workstations. All participants need to record their ID and write down their \textit{pathway} feedbacks for a given image. The MOS for the image and possible low-level attributes are listed as reference. A screenshot of the interface is shown in Fig.~\ref{fig:interface}.

\subsection{Prompts for Building Q-Instruct with GPT}

\paragraph{\textit{What/How} questions.} \textit{Generate multiple question and answer pairs based on the following description of an image quality. The questions can start with "What/Why/How". The answer should be concise and only contain the core information with minimum words. You should also generate several false answers for each question under the key of “false candidates”, which are also reasonable given the question by contradicts with the description. Organize the output a list in JSON format and when you respond, please only output the json, no other words are needed:}

\noindent\textit{Description: \$DESC}

\paragraph{\textit{Yes/No} questions.} \textit{Generate multiple yes-or-no question and answer pairs based on the following description of an image quality. The answer should be concise and only contain "Yes" or "No". The number of questions with the answer "Yes" should be close to the number of questions with the answer "No". You can also ask questions about quality issues that are not mentioned in the analysis. The answer for those unsure questions should be "No". Organize the output a list in JSON format and when you respond, please only output the json, no other words are needed:}

\noindent\textit{Description: \$DESC}

\paragraph{\textit{Extended} conversations.}  \textit{Generate conversations based on the following description of quality and other low-level visual attributes of an image. These conversations can include one of the aspects in the folow 1. Examining the causes of low-level visual patterns; 2. Providing improvement suggestions on photography; 3. Providing tools to restore, enhance, or edit the image; 4. Recommending the image to respective consumers; 5. Other conversations that may happen given the descriptions. Remember to be relevant to the image. Organize the output a list in JSON format (interleaved with "query" and "response" keys for each conversation) and when you respond, please only output the json, no other words are needed:}

\noindent\textit{Description: \$DESC}

\section{Hyper-parameters during Training}
\label{sec:supphyper}

\paragraph{Hyper-parameters for LLaVA-v1.5.}
The \textit{low-level visual instruction tuning} for LLaVA-v1.5 (7B/13B) is conducted with 8 NVIDIA A100-SMX4-80GB GPU (\textit{requiring 16 hours for 7B, 22 hours for 13B}, for the \textbf{\textit{mix}} version). We record all hyper-parameters in Tab.~\ref{tab:hyperparamllava}.

\begin{table}[htbp]
\centering
\resizebox{\linewidth}{!}{
\begin{tabular}{l| c c}
\toprule
Hyper-parameter &  \textbf{\textit{mix}} with high-level &  \textbf{\textit{after}} high-level \\
\midrule
ViT init. & \multicolumn{2}{c}{CLIP-L/14-336 \cite{CLIP}}   \\
LLM init. & Vicuna-v1.5~\cite{vicuna} & LLaVA-v1.5 \\
image resolution         & $336\times 336$ & $336\times 336$ \\
group modality length & True & False \\
batch size & \multicolumn{2}{c}{128} \\
lr max & \multicolumn{2}{c}{2e-5} \\
lr schedule & \multicolumn{2}{c}{cosine decay} \\
warmup epochs & \multicolumn{2}{c}{0.03} \\
weight decay & \multicolumn{2}{c}{0} \\
gradient acc.            & \multicolumn{2}{c}{1} \\
numerical precision      & \multicolumn{2}{c}{$\mathtt{bfloat16}$} \\
epoch & \multicolumn{2}{c}{1} \\
optimizer & \multicolumn{2}{c}{AdamW} \\
optimizer sharding       & \multicolumn{2}{c}{\checkmark} \\
activation checkpointing & \multicolumn{2}{c}{\checkmark} \\
deepspeed stage & \multicolumn{2}{c}{3} \\
\bottomrule
\end{tabular}
}
% \vspace{1mm}
\caption{
\textbf{Hyper-parameters} of \textit{low-level visual instruction tuning} on LLaVA-v1.5 (7B/13B), the same as original LLaVA-v1.5.
}
\label{tab:hyperparamllava}
\vspace{-10pt}
\end{table}

\paragraph{Hyper-parameters for mPLUG-Owl-2.} The \textit{low-level visual instruction tuning} for mPLUG-Owl-2 is conducted with 32 NVIDIA A100-SMX4-80GB GPU (requiring \textit{8 hours} for the \textbf{\textit{mix}} version). Hyper-parameters in Tab.~\ref{tab:hyperparammplugowl}.

\begin{table}[htbp]
    \centering
    %\tablestyle{7pt}{1.0}
    \resizebox{\linewidth}{!}{\begin{tabular}{l|cc}
         \toprule
Hyper-parameter            & \textbf{\textit{mix}} with high-level &  \textbf{\textit{after}} high-level \\
         \midrule
         ViT init.                & \multicolumn{2}{c}{Pre-train stage (updated CLIP-L/14 \cite{CLIP})} \\
         LLM init.                & \multicolumn{2}{c}{LLaMA-2 \cite{llama2}} \\
         visual abstractor init.  & Pre-train stage & mPLUG-Owl-2 \\
         image resolution         & $448\times 448$ & $448\times 448$ \\
         batch size & \multicolumn{2}{c}{256} \\
        lr max & \multicolumn{2}{c}{2e-5} \\
lr schedule & \multicolumn{2}{c}{cosine decay} \\
lr warmup ratio & \multicolumn{2}{c}{0.03} \\
weight decay & \multicolumn{2}{c}{0} \\
gradient acc.            & \multicolumn{2}{c}{16} \\
numerical precision      & \multicolumn{2}{c}{$\mathtt{bfloat16}$} \\
epoch & \multicolumn{2}{c}{1} \\
warm-up steps            & \multicolumn{2}{c}{250} \\
optimizer & \multicolumn{2}{c}{AdamW} \\
optimizer sharding       & \multicolumn{2}{c}{\checkmark} \\
activation checkpointing & \multicolumn{2}{c}{\checkmark} \\
model parallelism        & \multicolumn{2}{c}{2}\\
pipeline parallelism     & \multicolumn{2}{c}{1} \\
         \bottomrule
    \end{tabular}}
    \caption{\textbf{Hyper-parameters} of \textit{low-level visual instruction tuning} on mPLUG-Owl-2, the same as the original model.}
    \label{tab:hyperparammplugowl}
    \vspace{-10pt}
\end{table}

\paragraph{Hyper-parameters for InternLM-XComposer-VL.} Similar as mPLUG-Owl-2, the \textit{low-level visual instruction tuning} for InternLM-XComposer-VL is conducted with 32 NVIDIA A100-SMX4-80GB GPU (requiring \textit{13 hours} for the \textbf{\textit{mix}} version). Hyper-parameters are listed in Tab.~\ref{tab:hyperparamxcomposer}.

\begin{table}[htbp]
    \centering
    %\tablestyle{7pt}{1.0}
    \resizebox{\linewidth}{!}{\begin{tabular}{l|cc}
         \toprule
Hyper-parameter            & \textbf{\textit{mix}} with high-level &  \textbf{\textit{after}} high-level \\
         \midrule
         ViT init.                & \multicolumn{2}{c}{EVA-CLIP-G \cite{evaclip}} \\
         LLM init.                & Pre-train stage & InternLM-XComposer-VL \\
         perceive sampler init.  & Pre-train stage & InternLM-XComposer-VL \\
         image resolution         & $224\times 224$ & $224\times 224$ \\
         batch size & \multicolumn{2}{c}{256} \\
        lr max & \multicolumn{2}{c}{5e-5} \\
lr schedule & \multicolumn{2}{c}{cosine decay} \\
lr warmup ratio & \multicolumn{2}{c}{0.05} \\
weight decay & \multicolumn{2}{c}{0} \\
gradient acc.            & \multicolumn{2}{c}{1} \\
numerical precision      & \multicolumn{2}{c}{$\mathtt{float16}$} \\
epoch & \multicolumn{2}{c}{1} \\
warm-up steps            & \multicolumn{2}{c}{250} \\
optimizer & \multicolumn{2}{c}{AdamW} \\
special setting       & \multicolumn{2}{c}{low-rank adaptation (\textit{LORA})} \\
activation checkpointing & \multicolumn{2}{c}{\checkmark} \\
         \bottomrule
    \end{tabular}}
    \caption{\textbf{Hyper-parameters} of \textit{low-level visual instruction tuning} on InternLM-XComposer-VL, the same as the original model.}
    \label{tab:hyperparamxcomposer}
    \vspace{-10pt}
\end{table}
\section{Evaluation Details}

\subsection{Prompt Settings on (A1) Perception (\textit{via} MCQ)}

Denote the image tokens as {\tt <image>}, the question as {\tt <QUESTION>}, choices as {\tt <CHOICE$_i$>}, the prompt settings for different models on answering Multi-Choice Questions (MCQ) are slightly different, listed as follows. To ensure optimal results, during training, we also transform the VQA subset under the same settings, respectively.

\begin{figure*}
    \centering
    \includegraphics[width=0.7\textwidth]{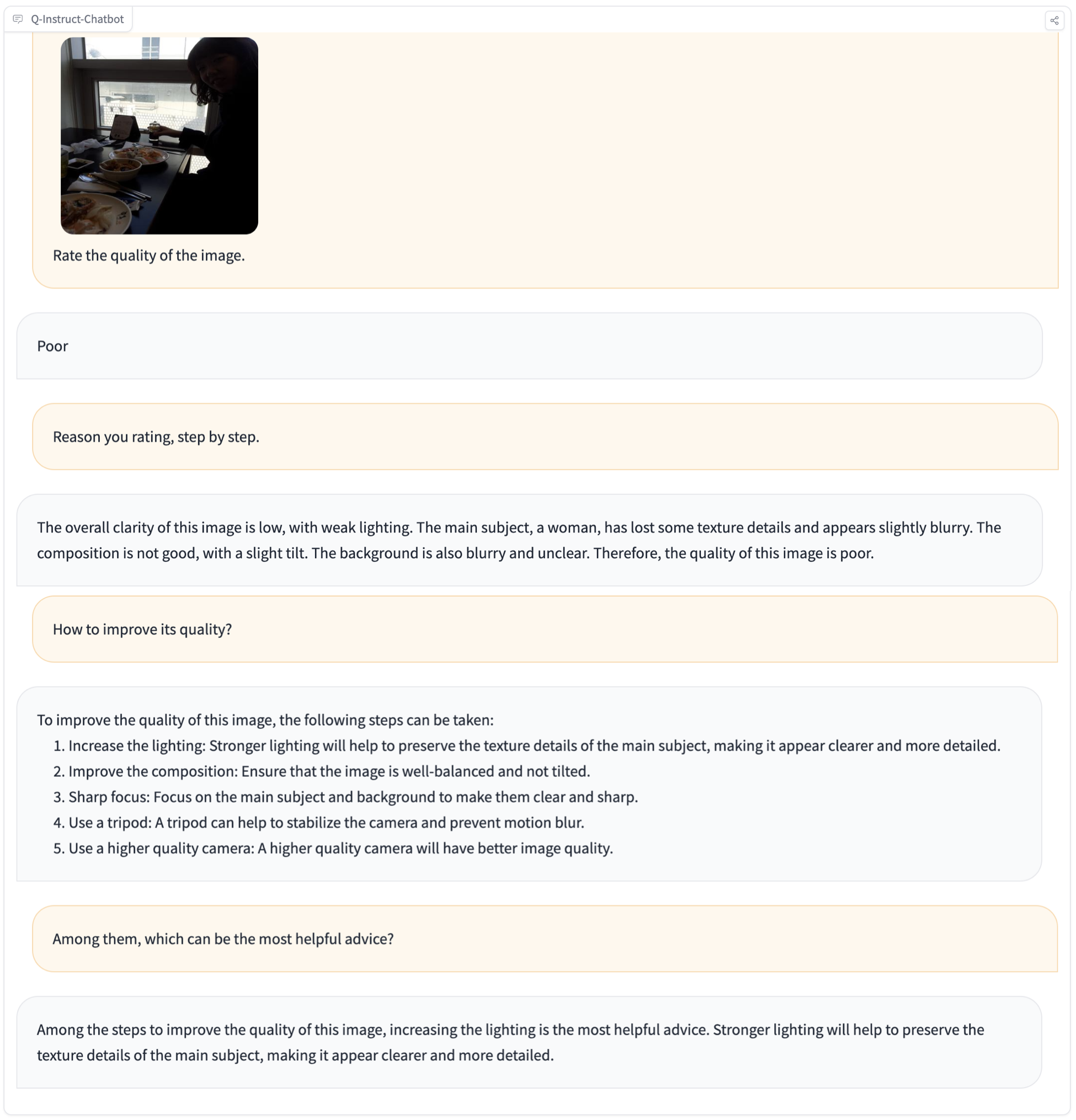}
    \vspace{-8pt}
    \caption{\textbf{Qualitative Analysis (I)}: A multi-turn conversation that the user subsequently queries the \textbf{Q-Instruct}-\textit{tuned} MLLM on (1) \textit{rating image quality}, (2) \textit{reasoning the rating}, (3) \textit{providing improvement suggestions}, and (4) \textit{discerning the most important suggestion}. }
    \label{fig:vi1}
    \vspace{-10pt}
\end{figure*}

\begin{figure*}
    \centering
    \includegraphics[width=0.73\textwidth]{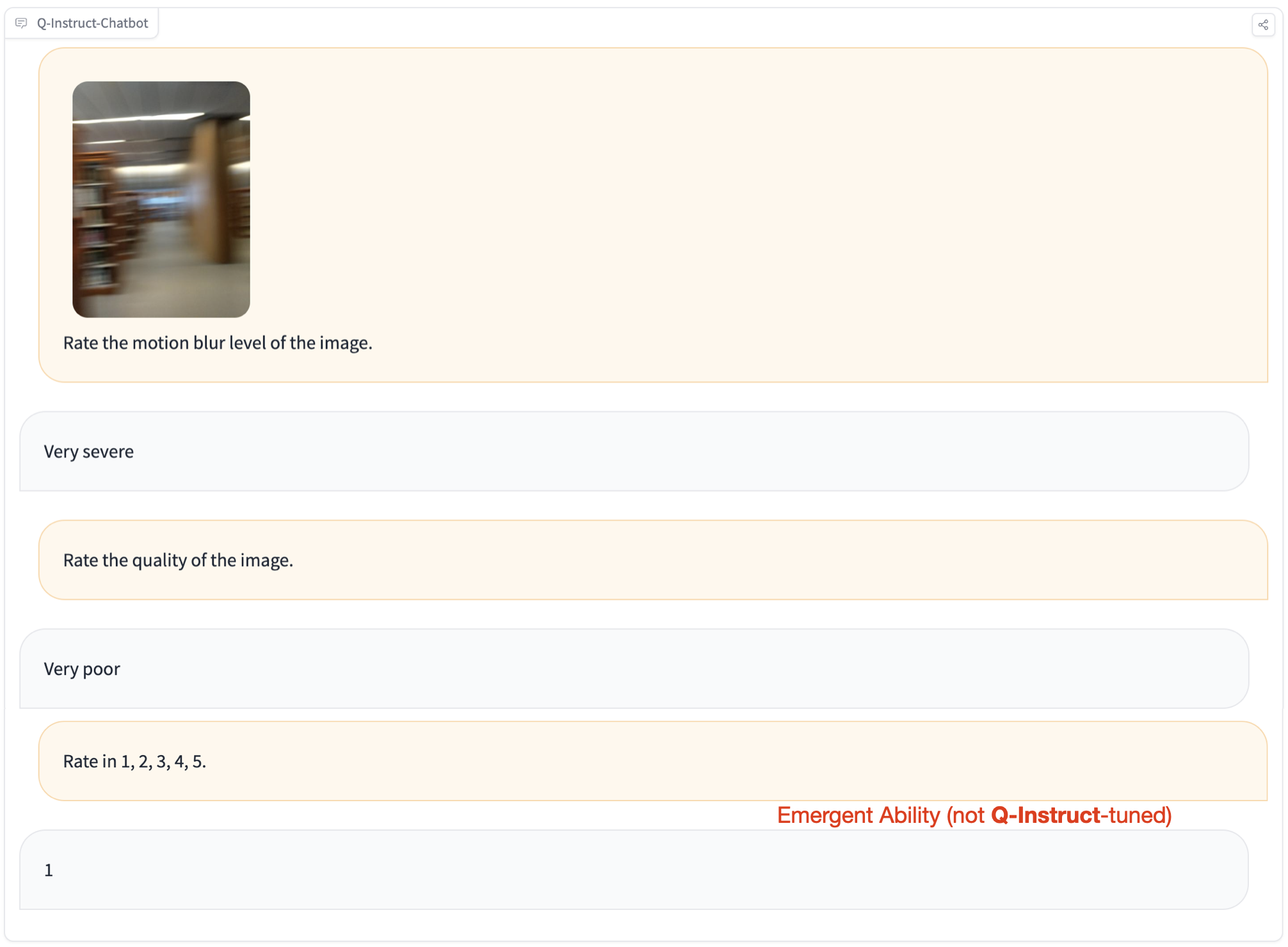}
    \vspace{-8pt}
    \caption{\textbf{Qualitative Analysis (II)}: The \textbf{Q-Instruct}-\textit{tuned} MLLM can serve as a multi-purpose (\textit{overall quality} or \textit{specific distortion}) and multi-format (\textit{text, good/average/poor} or \textit{numerical, e.g. 1/2/3/4/5}) quality evaluator. }
    \label{fig:vi2}
    \vspace{-8pt}
\end{figure*}

\paragraph{Prompt Settings for LLaVA-v1.5 (7B/13B).} \textit{A chat between a curious human and an artificial intelligence assistant. The assistant gives helpful, detailed, and polite answers to the human's questions. USER:}{\tt<image>} \\
\noindent{\tt <QUESTION>} \\
\textit{Answer with the option's letter from the given choices directly.} \\
\textit{A.} {\tt <CHOICE$_A$>} \\
\textit{B.} {\tt <CHOICE$_B$>} \\
\textit{C.} {\tt <CHOICE$_C$>} \\
\textit{ASSISTANT:}

\paragraph{Prompt Settings for mPLUG-Owl-2.} \textit{USER: {\tt<image>} \\ \noindent{\tt <QUESTION>}} \\
\textit{Answer with the option's letter from the given choices directly.} \\
\textit{A.} {\tt <CHOICE$_A$>} \\
\textit{B.} {\tt <CHOICE$_B$>} \\
\textit{C.} {\tt <CHOICE$_C$>} \\
\textit{ASSISTANT:}

\paragraph{Prompt Settings for InternLM-XComposer-VL.}\\
{\tt <|User|>}: {\tt<image>}\textit{Please answer this question by choosing the correct choice.Context: N/A} \\
\textit{Question:} {\tt <QUESTION>} \\
\textit{Options:} \textit{A.} {\tt <CHOICE$_A$>} \\
\textit{B.} {\tt <CHOICE$_B$>} \\
\textit{C.} {\tt <CHOICE$_C$>} \\
{\tt <TOKENS\_UNUSED\_0>} {\tt <|Bot|>}: \textit{Answer: The answer is}

\subsection{Prompt Setting on (A2) Description}

For the \textbf{(A2) Description} task, we unify all models under the same prompt: \textit{``Describe and evaluate the quality of the image."}, as this is the only prompt that can effectively allow every base model to describe low-level visual attributes and then evaluate image quality. For the alternate prompt as shown in Fig.~\ref{fig:one}, \textit{``Rate the quality of the image. Think step by step."}, the base InternLM-XComposer-VL only provides numbers (\textit{1/2/3/4/5}) without explanations or reasonings. Therefore, we choose the current prompt to evaluate the description ability among all variants.

\subsection{Prompt Setting on (A3) Assessment}

For the \textbf{(A3) Quality Assessment} task, we follow the strategy as proposed by Q-Bench~\cite{wu2023qbench}, with the {\tt softmax} output between \textit{good} and \textit{poor} to collect better \textit{quantifiable} scores for images, under the first output token of MLLMs:
\begin{equation}
q_\mathrm{pred} = \frac{e^{x^\text{\textbf{good}}_{\textit{SCORE\_TOKEN}}}}{e^{x^\text{\textbf{good}}_{\textit{SCORE\_TOKEN}}}+e^{x^\text{\textbf{poor}}_{\textit{SCORE\_TOKEN}}}}
\label{eq:1}
\end{equation}

For KoNViD-1k, the video quality assessment dataset as evaluated, we sample \textit{one frame per second} (the same as NIQE~\cite{niqe}), and average the image quality scores as calculated by Eq.~\ref{eq:1} among frames. While we acknowledge that this is not the optimal strategy for evaluating video quality, the excellent results in Tab.~\ref{tab:assessment} already significantly outperform all methods that are not trained for video quality assessment, suggesting that if better strategies are adopted, MLLMs can do even better on video quality assessment.

\section{Qualitative Analyses}
\label{sec:qualitative}

\begin{figure*}
    \centering
    \includegraphics[width=0.73\linewidth]{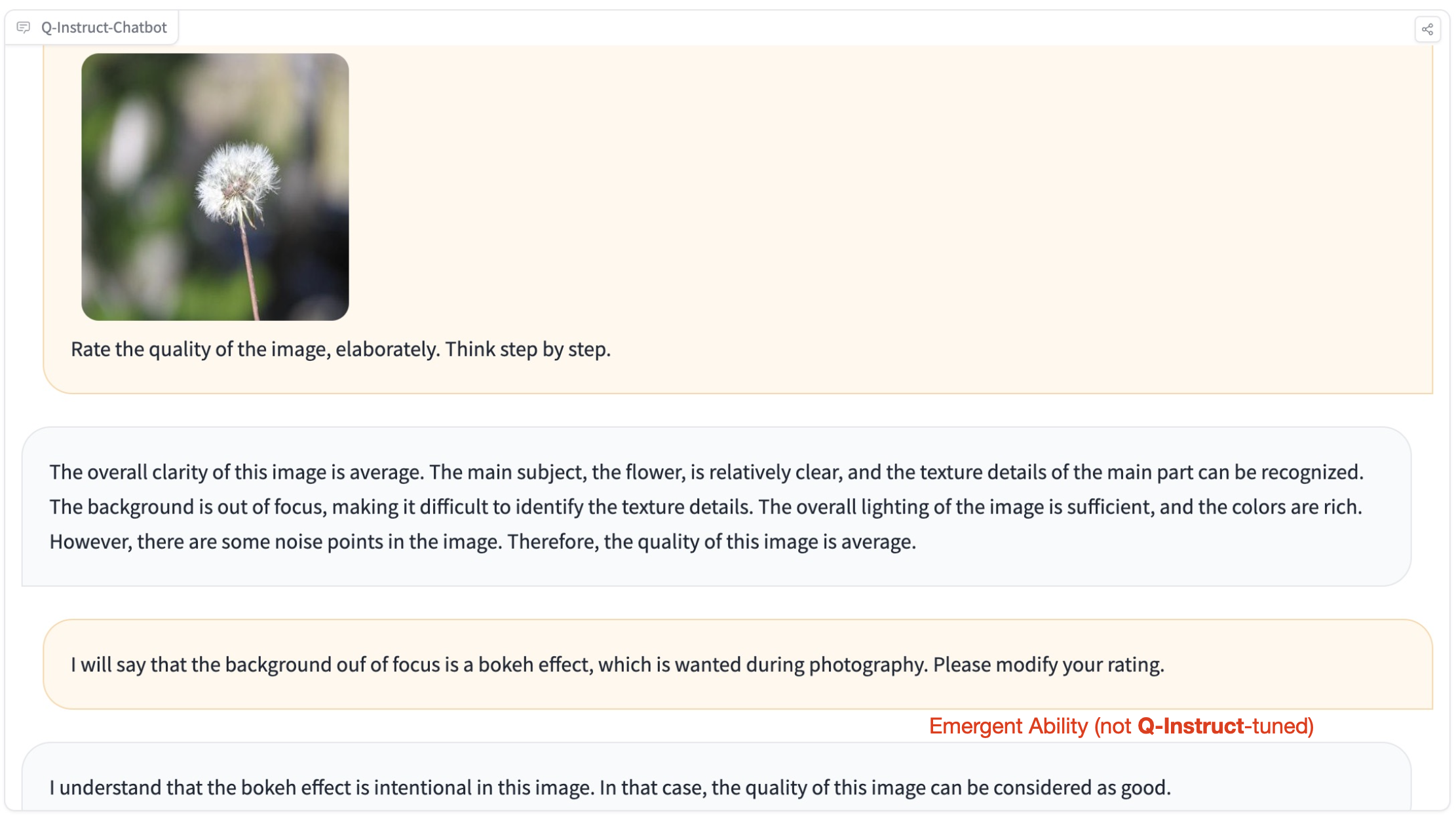}
    \vspace{-8pt}
    \caption{\textbf{Qualitative Analysis (III)}: The \textbf{Q-Instruct}-\textit{tuned} MLLM can correct itself based on further instructions. While provided with additional context (\textit{i.e.} background bokeh is intentional), it can modify its rating to align with the context.}
    \label{fig:vi3}
    \vspace{-10pt}
\end{figure*}

\begin{figure*}[ht!]
    \begin{subfigure}[b]{\textwidth}
    \centering
        \includegraphics[width=0.7\textwidth]{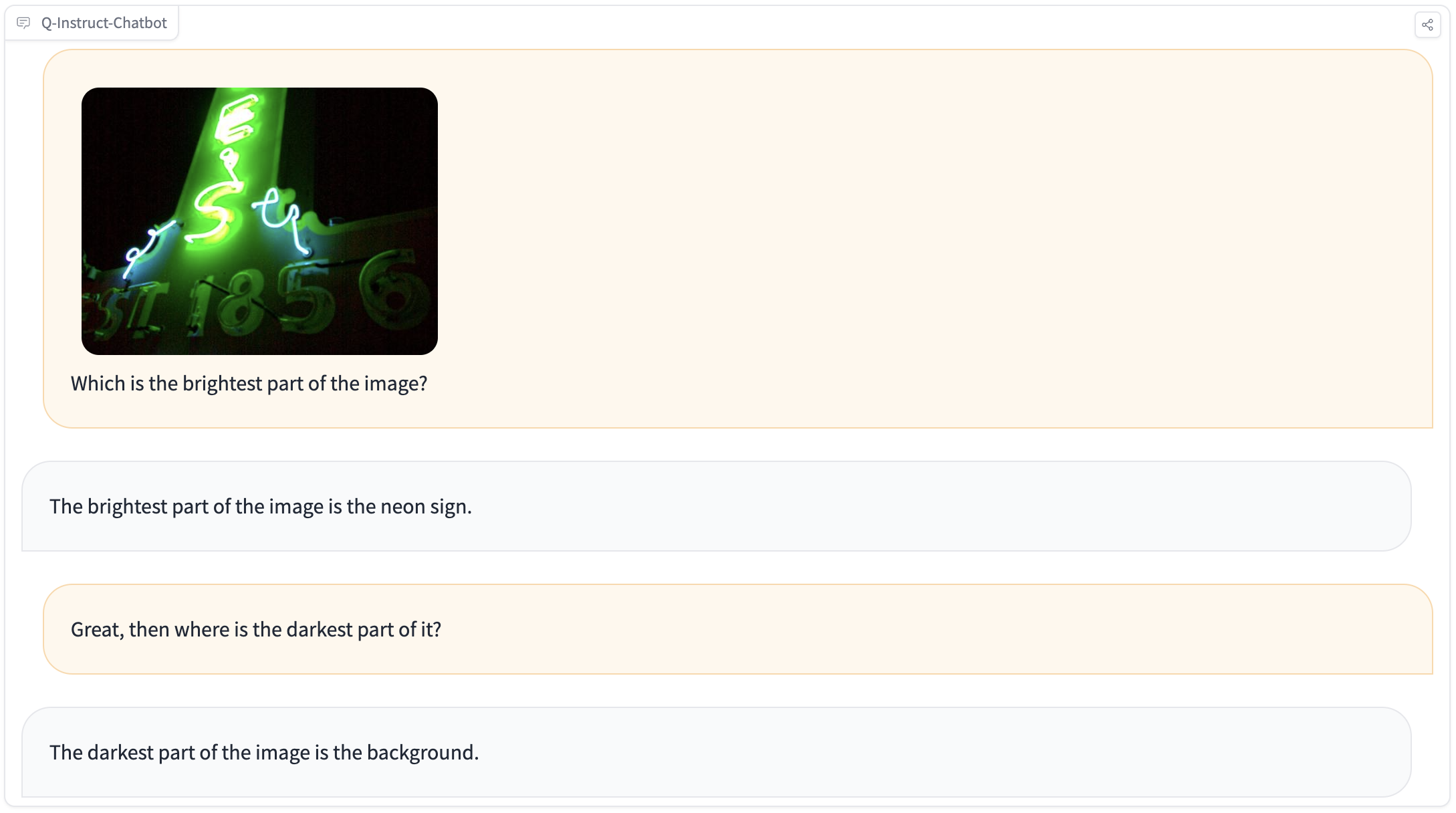}
        \caption{\textit{A strong contrast image.}}
        \label{fig:subfig1}
    \end{subfigure}%
    ~ %add desired spacing between images, e. g. ~, \quad, \qquad, \hfill etc. 
      %(or a blank line to force the subfigure onto a new line)

    \begin{subfigure}[b]{\textwidth}
        \centering

        \includegraphics[width=0.7\textwidth]{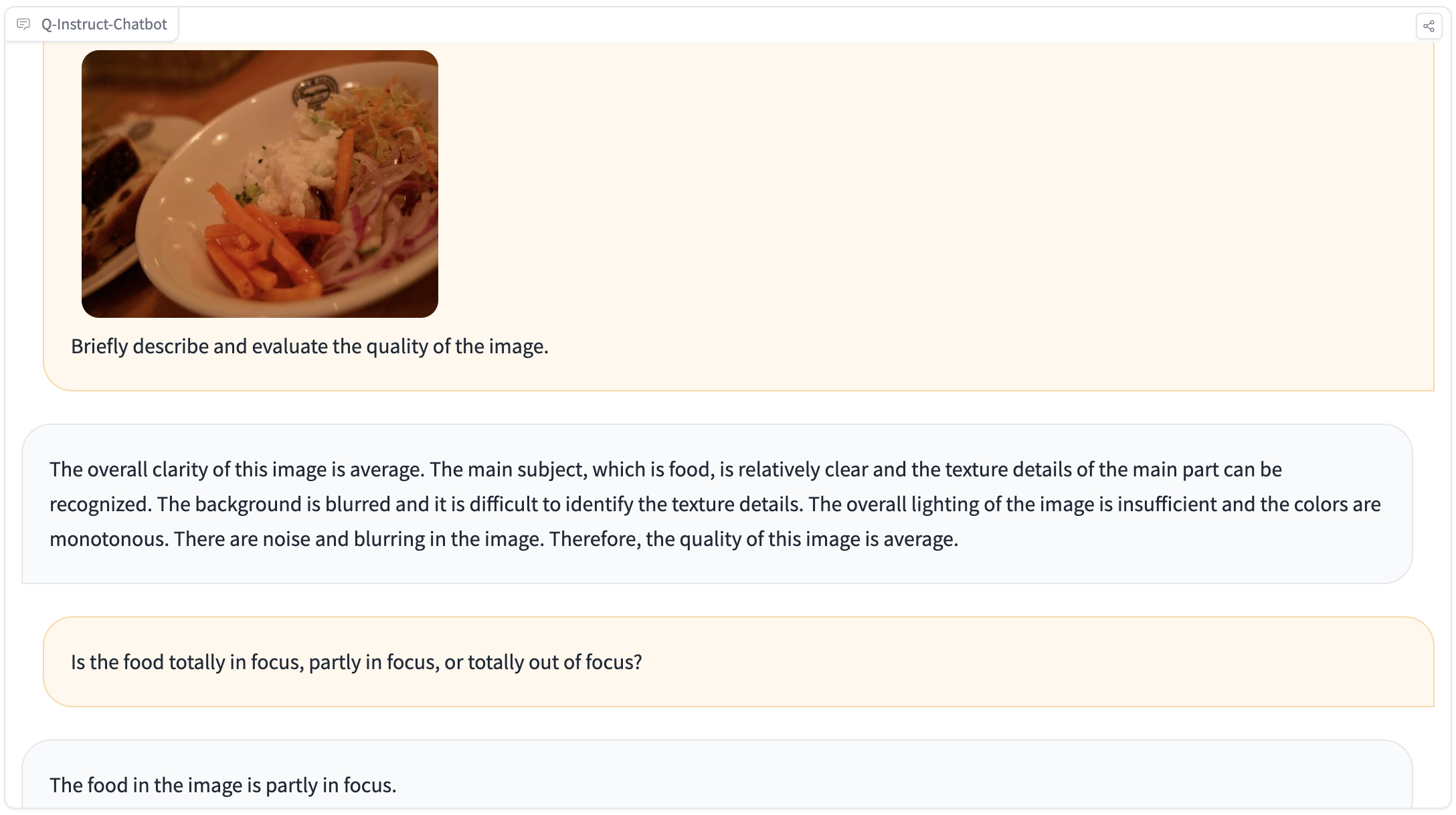}
        \caption{\textit{A partly in-focus image.}}
        \label{fig:subfig2}
    \end{subfigure}
    \caption{\textbf{Qualitative Analysis (IV)}: Local in-context low-level perceptual abilities of \textbf{Q-Instruct}-\textit{tuned} MLLMs. They can effectively discern the bright part and dark part in a \textit{strong contrast image} (a), or the clarity of different objects/areas in a \textit{partly in-focus image} (b).}
    \label{fig:vi4}
\end{figure*}

Considering that different base MLLMs reach similar performance after \textit{low-level visual instruction tuning}, we build a publicly-accessible demo for \textbf{Q-Instruct}-mPLUG-Owl-2 under the \textit{\textbf{mix}} strategy, and invite our team members to really \textit{chat} with it. These conversations exhibit its exellent abilities on various scenraios, including multi-turn queries (Fig.~\ref{fig:vi1}), multi-purpose assessment (Fig.~\ref{fig:vi2}), correction (Fig.~\ref{fig:vi3}), and local in-context perception  (Fig.~\ref{fig:vi4}). 

%\subsection{Low-level Visual Description}

%\subsection{Open-ended Conversation}

\section{Limitations}
\label{sec:limitation}

The known limitations of our studies are as follows. First, though with improved quality assessment and low-level visual perception abilities, the \textbf{Q-Instruct}-\textit{tuned} models have witnessed declined performance on general-purpose tasks, especially language-centric tasks, or tasks that require heavy reasoning abilities. Therefore, they may produce unwanted outputs if applied to tasks other than low-level visual perception and understanding. Second, though with improved accuracy, the \textbf{Q-Instruct}-\textit{tuned} models still perform worse (68\%-71\% accuracy on LLVisionQA-\textit{test}) than an average human (about 74\%), and may not yet be able to directly replace human on low-level related tasks. Thirdly, the \textbf{Q-Instruct} dataset mainly consists of natural in-the-wild images. Though they prove excellent generalization on other types of visual contents, the performance might still be improveable if further tuned on these datasets.

\section{Ethical Acknowledgements}
\label{sec:ethical}

In our study, all participants were fully informed about the nature and amount of the tasks involved prior to their participation. No uncomfortable content was reported during this process. We express our gratitude to the participants for their valuable contributions, which were essential to the success of our research. We commit to upholding all ethical standards to ensure the well-being of our participants, as well as the integrity of our research findings.

\section{Acknowledgements}

Our team would like to sincerely thank Qinghao Ye and Xiaoyi Dong for providing pre-trained weights of mPLUG-Owl-2 and InternLM-XComposer-VL after the \textit{low-level visual instruction tuning}, including a complete training recipe on their respective high-level datasets mixed with the \textbf{Q-Instruct}. We believe these weights will significantly contribute to the open-source community working on tasks related to low-level visual perception and understanding.

\section{License}

Researchers and open-source developers are free to use the \textbf{Q-Instruct} dataset and the fine-tuned weights as provided for the four MLLMs. We also allow commercial use, while any commercial use should be pre-permitted by our team. Any usage should also comply with licenses of the original base models (\textit{inc.} base LLMs such as Vicuna, LLaMA-2).

\begin{comment}
{
    \small
    \bibliographystyle{ieeenat_fullname}
    \bibliography{main}
}
\end{comment}

\end{document}